\documentclass[twoside,11pt]{article}

\usepackage{jmlr2e}
\usepackage{lastpage}
\jmlrheading{23}{2022}{1-\pageref{LastPage}}{10/20; Revised 6/22}{7/22}{20-1121}{Aidan N. Gomez, Oscar Key, Kuba Perlin, Stephen Gou, Nick Frosst, Jeff Dean, Yarin Gal}

\usepackage[utf8]{inputenc} 
\usepackage[T1]{fontenc}    
\usepackage{hyperref}       
\usepackage{url}            
\usepackage{booktabs}       
\usepackage{colortbl}       
\usepackage{multirow}
\usepackage{tabularx}
\usepackage{amsfonts}       
\usepackage{nicefrac}       
\usepackage{microtype}      
\usepackage{mathtools}      
\usepackage{xspace}
\usepackage{tikz}
\usetikzlibrary{calc}
\usepackage{natbib}
\usepackage{cleveref}
\usepackage{appendix}
\usepackage{comment}
\usepackage{layouts}
\usepackage{natbib}
\usepackage{xr-hyper}
\usepackage{xspace}
\usepackage{enumitem}

\newcommand{\etoe}{\textsc{end-to-end}\xspace}
\newcommand{\local}{\textsc{1-wise}\xspace}
\newcommand{\nwise}[1]{\textsc{#1-wise}}
\newcommand{\finetuning}{\textsc{fine-tuning}\xspace}

\newcommand{\hicell}{\cellcolor{gray!20}} 

\title{Interlocking Backpropagation: Improving depthwise model-parallelism}

\ShortHeadings{Interlocking Backpropagation}{Gomez, Key, Perlin, Gou, Frosst, Dean, and Gal}
\firstpageno{1}

\begin{document}

\author{%
    \name Aidan N. Gomez\thanks{Joint first author} \email aidan.gomez@cs.ox.ac.uk \\
    \addr University of Oxford \& Cohere \\
    \AND
    \name Oscar Key\footnotemark[1] \email oscar.key.20@ucl.ac.uk \\
    \addr University of Oxford\\
    \AND
    \name Kuba Perlin \email kuba@cohere.ai \\
    \addr Cohere\\
    \AND
    \name Stephen Gou \email stephen@cohere.ai \\
    \addr Cohere\\
    \AND
    \name Nick Frosst \email nick@cohere.ai \\
    \addr Cohere\\
    \AND
    \name Jeff Dean \email jeff@google.com \\
    \addr Google\\
    \AND
    \name Yarin Gal \email yarin@cs.ox.ac.uk \\
    \addr University of Oxford\\
}

\editor{Sathiya Keerthi}
\maketitle

\begin{abstract}
The number of parameters in state of the art neural networks has drastically increased in recent years. This surge of interest in large scale neural networks has motivated the development of new distributed training strategies enabling such models. One such strategy is model-parallel distributed training. Unfortunately, model-parallelism can suffer from poor resource utilisation, which leads to wasted resources. In this work, we improve upon recent developments in an idealised model-parallel optimisation setting: local learning. Motivated by poor resource utilisation in the global setting and poor task performance in the local setting, we introduce a class of intermediary strategies between local and global learning referred to as \emph{interlocking backpropagation}. These strategies preserve many of the compute-efficiency advantages of local optimisation, while recovering much of the task performance achieved by global optimisation. We assess our strategies on both image classification ResNets and Transformer language models, finding that our strategy consistently out-performs local learning in terms of task performance, and out-performs global learning in training efficiency.
\end{abstract}


\section{Introduction}

Modern state-of-the-art language models require billions of parameters. These models are often too large to fit in the memory of a single accelerator, and so the training computation must be distributed across multiple accelerator devices. Training such large models can be accomplished by partitioning the model across several accelerators and communicating the activations and gradients between them. However, this naive approach to training way incurs significant inefficiencies, as each accelerator must wait for all downstream accelerators to compute their forwards and backwards passes before it can begin computation of its own backwards pass. This optimisation setting is referred to as `global learning', as there is a single global objective that must be evaluated in order to compute updates to the parameters.

An idealised model-parallel optimisation setting would be one where each accelerator need only push data to the next, never waiting for any returning gradient. In order to facilitate this, each accelerator's portion of the model must be able to compute weight updates with which to train itself, without access to any information from downstream accelerators. This idealised setting is referred to as `local learning' and has seen an uptick in recent interest \citep{etete,dgl}. However, there remain core limitations to the proposed methods, principal among them: the degradation in modelling performance relative to global learning.

In this work we attempt to improve the efficiency of distributed model training by exploring strategies that strike a middle-ground between local and global learning via backpropagation. We investigate a class of training regimes by training large-scale neural networks with auxiliary classification layers throughout the network, and restricting the gradient flow from each of these classification heads. We refer to these strategies as \textit{interlocking backpropagation}. We find that interlocking backpropagation is significantly more compute efficient than the standard global backpropagation approach, yet it recovers much of its modelling performance, compared to local learning.

Our work presents the following contributions:
\begin{itemize}
    \item We explore modelling limitations of local optimisation.
    \item We propose a class of optimisation algorithms that aim to \emph{preserve much of the compute efficiency} of local training, while \emph{significantly improving modelling performance}.
    \item We provide a generic, open-source framework for the study of this class of optimisation algorithms.
    It is available at \\ \url{https://github.com/oscarkey/interlocking-backprop}.
\end{itemize}

\section{Related Work}

Previous work has attempted to improve the resource utilisation of distributed model-parallel training regimes. GPipe \citep{gpipe} addresses the inefficiency of training model-parallel distributed networks by splitting \emph{mini}-batches into \emph{micro}-batches and processing these micro-batches concurrently. GPipe accumulates gradients of all micro-batches before applying them to the weights. That strategy increases resource utilisation, but it still requires more time overall than a local approach to training.

\begin{figure}
    \centering
    \includegraphics[width=0.55\textwidth, page=4]{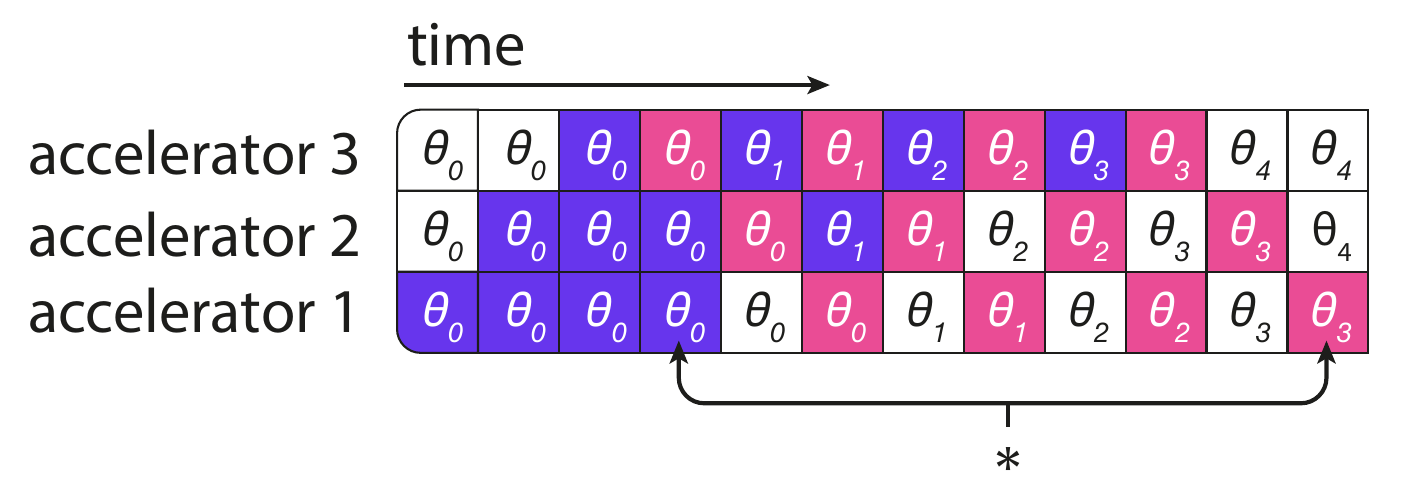}
    \vspace{-0.4cm}
    \caption{
        Depiction of the problem of stale gradients in Hogwild-style training.
        Here we have a model split across three accelerators.
        Each row shows the evolution of the parameters in the memory of a given accelerator, with each column representing a time step.
        \(\theta_i\) denotes the parameters after the $i^\text{th}$ gradient update.
        The steps highlighted in purple indicate that the accelerator performed a forward pass on a single batch, while those in pink indicate a backward pass.
        (\textasteriskcentered{}) During the forward pass of the fourth batch of data, the first accelerator computes its activations using the parameters \(\theta_0\) (left arrow); however, during the fourth batch's backwards pass the first module has already had its parameters updated three times to \(\theta_3\) (right arrow).
        This mismatch between the weights used to compute the activations and those used to compute the gradient can disrupt optimisation dramatically (see Table~\ref{tab:transformer-hogwild} in Appendix~\ref{app:additional_results}).
    }
    \label{fig:hogwild}
\end{figure}

While GPipe performs weight updates that are equivalent to global learning, there are strategies to speeding up distributed training that do not have this property. One such strategy is called Hogwild \citep{hogwild}. Instead of performing a forward pass on each accelerator and waiting until the full backwards pass of the network has been computed and communicated to update the parameters, Hogwild starts processing the forward pass of subsequent batches as soon as the first accelerator has completed the forward pass of the first batch. Likewise, as soon as an earlier accelerator receives gradients from the subsequent accelerator, Hogwild applies those gradients and updates the weights. This means that when the gradients for the second batch are applied, they are being applied to different weights than those that were used to compute the corresponding forward pass. This problem is referred to as \emph{stale gradients}, and has been shown to introduce training instabilities. It is visualised in Figure \ref{fig:hogwild}.
Hogwild overlooks the issue of stale gradients entirely, opting to \emph{greedily} apply gradients to the weights as soon as any arrive.
This results in a very fast-executing optimisation strategy, but it can have a dramatic impact on the stability of optimisation when many accelerators are involved, making the method less favourable.
The Hogwild approach is orthogonal to interlocking backpropagation, the method we introduce in this paper, as interlocking backpropagation could be run with greedy gradient updates as well.
This however would introduce the same issue of stale gradients.
In our experiments we find that Hogwild greatly underperforms both global learning and interlocking backpropagation (see Appendix~\ref{app:additional_results}, Table~\ref{tab:transformer-hogwild}).

Local learning resolves the stale gradient issue by doing away with the communication of gradients entirely.
Each accelerator is responsible for using its own training signal to compute and apply parameter updates.
This also means each accelerator spends no idle time whatsoever waiting for gradients from other accelerators -- which is why it is considered an idealised setting for distributed model optimisation.
Local objectives were initially used in early unsupervised methods for training neural networks, such as the wake-sleep algorithm \citep{wake-sleep}.
In Inception networks \citep{inception-nets}, local objectives were applied in addition to the overall global objective in order to solve the vanishing gradients problem, though this approach has largely been supplanted by residual connections \citep{he2016deep}.
More recently, several variations on the local learning paradigm have been introduced with a focus on parallel training.
While highly parallelisable, many of these approaches do not match the accuracy of globally trained models \citep{mostafa-local-learning, dni}.
Others do achieve performance comparable to that of global learning, by using specially crafted local objective functions.
\citet{etete} use a contrastive predictive loss to achieve excellent performance in an unsupervised setting.
\citet{predsim} consider supervised local learning, and succeed in matching the accuracy of global learning on classification tasks with up to $100$ classes.
However, the authors note that their \emph{predsim} loss is not suited for classification tasks with larger numbers of classes, due to the similarity matrix becoming sparse. The proposed work-around of limiting the number of classes in each batch was shown to be effective for CIFAR-100, but is not applicable to some tasks such as language modelling.
In contrast to the aforementioned approaches, our method is a high-level paradigm for interpolating between end-to-end and local training, applicable to both the unsupervised and supervised settings, and possible to combine with specialised loss functions, such as predsim.

One reason why some local learning algorithms fail to match the performance of global algorithms is that there is no mechanism for layers higher in the model to communicate with lower layers of the model.
This means that the lower layers may learn representations which are not productive for learning in the higher layers.
For example, a lower layer may discard information from which a higher layer could have extracted additional accuracy.
\citet{loco} attempt to solve this problem by allowing the gradient signal to flow between pairs of adjacent accelerators.
Importantly, the pairs overlap with each other, meaning that the upper accelerator of one pair is the lower accelerator of the subsequent pair, thus allowing indirect gradient flow through the entire model
(see Section~\ref{sec:method} for a more detailed description of this structure, as a special case of interlocking backpropagation).
Introduced concurrently to our work, this approach is similar to our \nwise{n} algorithm, detailed later, for the particular case where $N=2$.
Our more general framework allows the practitioner to trade off training speed for increased accuracy, and we provide substantial experimental results investigating this trade-off.
We also focus our experiments on supervised rather than unsupervised learning,
and we specifically test our algorithm on large Transformer models with hundreds of millions of parameters.

In further concurrent work, ~\citet{laskin-metz} also inspect interlocking backpropagation, referring to it as `overlapping local updates', and present a cost-performance analysis of \nwise{2}, \etoe, and alternative training strategies, applied to tasks in image and language domains.

\section{Interlocking Backpropagation}
\label{sec:method}

\begin{figure}[t]
    \centering
    \includegraphics[width=1\textwidth, page=2]{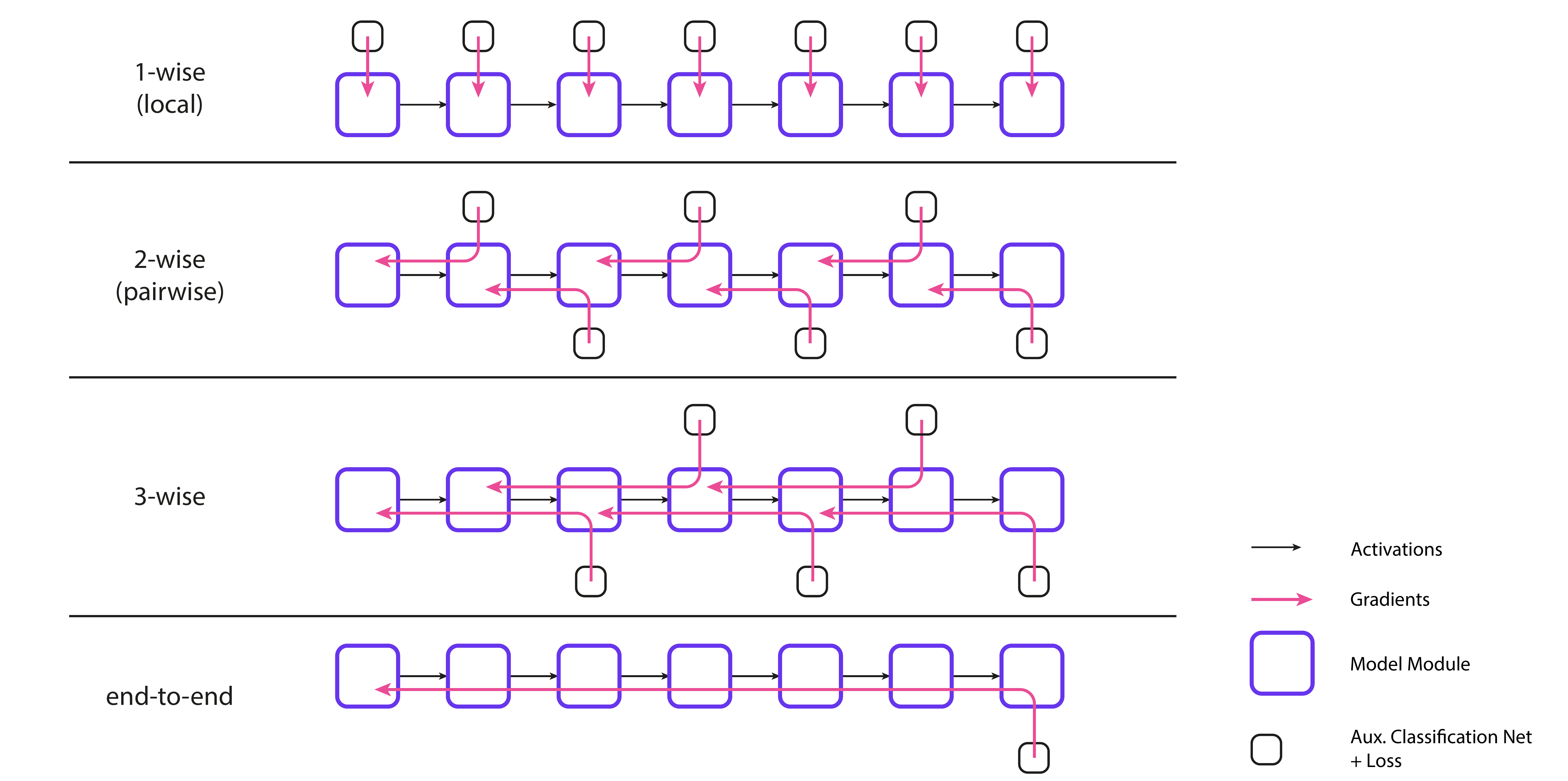}
    \caption{Depiction of the flow of activations and gradients through interlocking backpropagation for different optimisation strategies.
    Activation flows are shown in black and gradients are shown in red.
    One extreme is \nwise{1} optimisation (local learning), where there is no gradient communication between modules.
    The other extreme is \etoe optimisation, where gradients flow through all modules from a global loss function at the top of the network.
    The \nwise{2} and \nwise{3} strategies, as introduced in this paper, strike a middle ground.
    In \nwise{2}, gradients flow from an auxiliary network attached to each module, through the local module, and travel one module boundary before stopping.
    Similarly, \nwise{3} has gradients travel through two module boundaries before stopping.}
    \label{fig:gradient_flow}
\end{figure}

A neural network can be described as a composition of a series of smaller functions; for example \(f = f_6 \circ \dots \circ  f_2 \circ f_1\).
When the parameterisation of the network exceeds the limit of a single hardware accelerator, contiguous groups of these functions can be placed on individual accelerators.
Each of these contiguous groups is referred to as a \emph{module}.
Note that in this work we assume that each accelerator holds only one module, as having multiple modules on a single accelerator would introduce unnecessary overheads to training.
If an accelerator holds more than one module, these can simply be merged together to form a single module.

The communication between modules can be costly, and so one could attempt to speed up the learning process by performing local learning on each module.
Consider a network composed of three modules of two layers each:
\begin{align*}
f &= f_{c_3} \circ f_{c_2} \circ f_{c_1} \\
\text{where, }\,f_{c_1} &= f_2 \circ f_1, \text{ parameterised by } \theta_{c_1}=(\theta_2,\theta_1) \\
f_{c_2} &= f_4 \circ f_3, \text{ parameterised by } \theta_{c_2}=(\theta_4, \theta_3) \\
f_{c_3} &= f_6 \circ f_5, \text{ parameterised by } \theta_{c_3}=(\theta_6, \theta_5) .
\end{align*}
We consider several possible approaches for training this model, which differ in the amount of communication between modules. One extreme, involving the most communication, is \etoe training. Here we compute the loss based on the output of the final module, $f_{c_3}$, and propagate the loss backwards through each module to update their parameters. This approach is depicted in the bottom row of Figure \ref{fig:gradient_flow} and it achieves identical accuracy to if the model was in a single module on a single accelerator; however, the communication between modules during the backwards pass leads to inefficiencies. For instance, the first module in the model, having completed its forward pass, must sit idle, while it waits for the modules above it to complete their forward and backward passes. Only then, does it receive the gradient signal and is able to perform its own backwards pass.

The other extreme, resulting in the least idleness, is local training, illustrated in the first row of Figure \ref{fig:gradient_flow}.
In this setting, we augment each module with a local loss function, \(\mathcal{L}_{c_k}\):
\begin{align*}
    \mathcal{L}_{c_k}(x, y) &= \mathcal{L}(\hat{y}_{c_k}(x), y) \\
    \text{where, }\,\hat{y}_{c_k}(x) &= h_{c_k}(f_{c_k} \circ \dots \circ f_{c_1}(x)) .
\end{align*}
Here, \(x\) is the training input to the model, \(y\) is the target, and \(\mathcal{L}\) is a standard loss function, such as the cross-entropy loss.
We call \(h_{c_k}\) the auxiliary network for module \(k\). It produces predictions for the task directly from the outputs of the $k^\text{th}$ module. During training, we update the parameters of both the main and auxiliary networks of the $k^\text{th}$ module based \emph{only} on gradients from \(\mathcal{L}_{c_k}\). This means that during the backwards pass no communication is required between modules. Thus, the only communication necessary between the hardware accelerators holding each module is to propagate the activations during the forward pass. This strategy avoids accelerators idling during the backward pass, while they wait to receive gradients from subsequent accelerators.

\begin{figure}
    \centering
    \vbox{
    \begin{minipage}[ht]{1\textwidth}
        \centering
        \includegraphics[width=\textwidth, page=1]{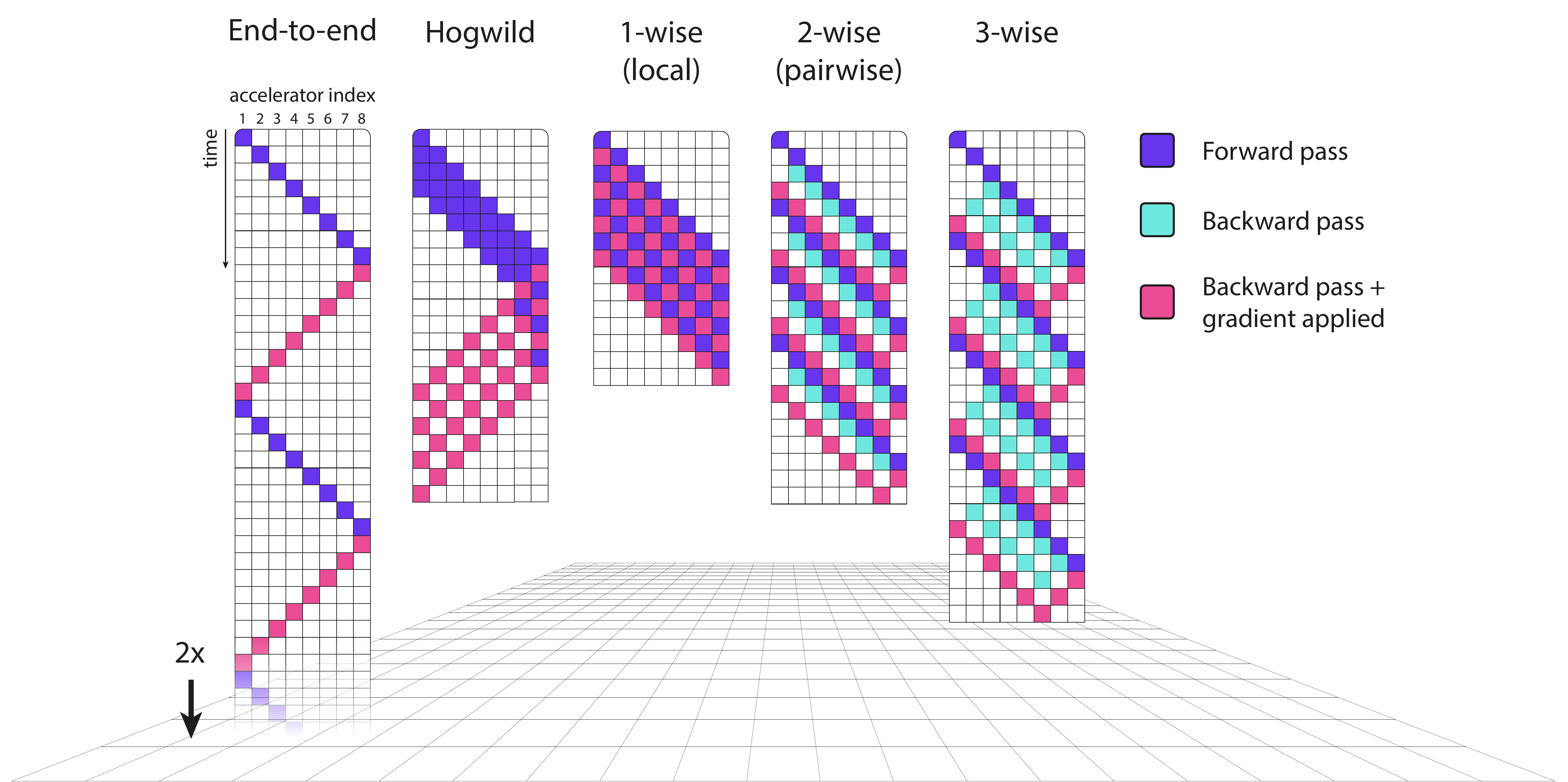}
    \end{minipage}
    }
    \vspace{0.25cm}
    \begin{tabular}{c c c c c c}
        \toprule
        \textbf{\etoe} & \textbf{\textsc{hogwild}} & \textbf{\nwise{1} (local)} & \textbf{\nwise{2}} & \textbf{\nwise{3}} & \textbf{\nwise{n}} \\
        \midrule
        $2A$ & $2$ & $2$ & $4$ & $6$ & $2N$ \\
        \bottomrule
    \end{tabular} \\
    \vspace{0.25cm}
    \caption{
    (top) Four training steps of different distributed optimisation strategies.
    Each row represents a time step, and each column an accelerator.
    \nwise{2} and \nwise{3} interlocking backpropagation, as introduced in this paper, are far cheaper than \etoe in terms of total optimisation time, and offer a natural trade off between speed and optimisation performance.\hspace{\textwidth}  
    (bottom) Table comparing the scaling of time per batch for different optimisation strategies applied to $A$ accelerators. Hogwild achieves its optimal time per batch when run long enough to fill its pipeline, which is not illustrated above.}
    \label{fig:cost-comparison}
\end{figure}

While local training is time efficient, without backwards communication between modules it fails to match \etoe in test accuracy. In this work we address this problem by introducing new intermediate strategies between \etoe and local training, where we allow varying amounts of communication between  modules. We refer to this family of strategies as \nwise{n} interlocking backpropagation.
Figure \ref{fig:gradient_flow} illustrates \nwise{2} and \nwise{3}.
Here the parameters in module $k$ are updated using gradients from \(\mathcal{L}_{c_{k+(N-1)}}\), which have been propagated backwards through the intermediate modules.
When \(N\) is set to \(1\), this is equivalent to local optimisation \citep{dgl} (i.e. \nwise{1}); when \(N\) is set to the number of modules, this is equivalent to global optimisation (i.e. \etoe).
By changing $N$, we can control the trade-off between performance and accuracy to match our application.
A slight variation on this approach is to update the parameters of module $k$ using the mean of the gradients propagated from both the \(\mathcal{L}_{c_{k}}\) and \(\mathcal{L}_{c_{k+(N-1)}}\).
We use this variant in our Transformer experiments as we find that, in this specific case, it improves modelling performance without affecting step time.

The step times of these strategies are visualised in Figure \ref{fig:cost-comparison}, which demonstrates that \nwise{2} and \nwise{3} are substantially faster than \etoe training.
In the next section we expand on this figure by presenting a detailed analysis of the step time of each training method.
While \nwise{n} has a time per gradient step that is always at least as fast as \etoe, we note that \nwise{n} will use more total computation because each module in a model trained using \nwise{n} must perform $N$ backward passes, while \etoe performs only one backward pass per module.
These additional backward passes do not cause any increase in training time because they occur while the accelerator would otherwise be idle, however it is possible they could affect the total energy used during training.
Whether \nwise{n} will increase or decrease total energy usage will depend on the exact scenario, as it is influenced by the power used by the accelerators when idle, and any reduction in total training time due to \nwise{n} being faster than \etoe to achieve a given accuracy (as demonstrated in Section \ref{sec:experiments}).

During \nwise{n} testing, we only make predictions using the output of the final module, and this is what we use to compute test accuracy. An alternative approach would be to ensemble the predictions of the auxiliary network at each module. However, experimentally we find that this does not significantly increase performance, likely due to the modules being highly correlated, and in some cases earlier modules having much lower accuracy than later ones (see Table \ref{tab:ensembled-predictions} in Appendix \ref{app:additional_results}).

\subsection{Training Speed of Interlocking Backpropagation}
Interlocking backpropagation allows shorter training step times than end-to-end learning because the gradients do not need to pass through the entire network.
The step time can be controlled and reduced dramatically by lowering the $N$ parameter of \nwise{n}.
A different model parallelism technique is pipeline parallelism, as used by GPipe~\citep{gpipe}.
Here each mini-batch is split into multiple micro-batches, allowing multiple mini-batches to be processed simultaneously by different modules of the network.
In fact, interlocking backpropagation can be combined with micro-batching as done in GPipe to yield further reductions in training time.
In this section we propose a general algebraic model of \textit{time per batch} for synchronous model-distributed learning.
Our model uses three parameters to represent local learning, end-to-end learning, GPipe and interlocking backpropagation, allowing an in-depth comparison of the performance of these methods.
Additionally, it allows us to examine the combination of interlocking backpropagation and micro-batching.
We fit this model to our experiment results, thus the predictions it makes are supported by practical data from experiments rather than being purely theoretical.

The parameters of the model are as follows:
\begin{itemize}
    \setlength{\parskip}{0pt}
    \setlength{\itemsep}{0pt}
    \item $A$ -- the number of sequential accelerators (modules).
    \item $M$ -- the number of micro-batches per mini-batch.
    \item $N$ -- the \nwise{n} parameter, in the range $[1, A]$.
\end{itemize}
If $M=1$ then the model corresponds to the training methods without micro-batching described earlier in the paper.
$N=1$ corresponds to local learning, $N=A$ to end-to-end learning, and $1 < N < A$ to \nwise{n} interlocking backpropagation.
If $M \geq 2$ then micro-batching is enabled.
In this case $N=A$ corresponds to GPipe as described by \citet{gpipe}, and other values of $N$ correspond to pipeline parallised versions of \nwise{n}.

As we are modelling a synchronous setting, we split the time domain into segments of equal duration, as visualised in Figure~\ref{fig:cost-comparison}.
We denote the duration of a single time slot by $c(M)$, and refer to it as the micro-batch processing cost. Our model makes the simplifying assumption that a forward and backward pass are the same duration. Relaxation of this assumption is discussed in Appendix~\ref{app:timing_model}.

The time per mini-batch (equivalently, per one gradient update) can then be derived as:
\begin{equation*}
T(A,M,N) = \begin{cases}
    (2 + A-N) \cdot Mc(M) + 2(2N-A-1)\cdot c(M) & \text{(for } M > 2, A < 2N-1 \text{)} \\
    (N + 1) \cdot Mc(M) & \text{(for } M > 2, A \ge 2N-1 \text{)} \\
    (N + 1) \cdot Mc(M) & \text{(for } M = 2 \text{)} \\
    2N \cdot Mc(M) & \text{(for } M = 1 \text{)}
    \end{cases}
\end{equation*}
For end-to-end learning ($N=A$), the formula degenerates to $T(A,M) = 2(M+A-1)\cdot c(M)$.
We define the micro-batch processing cost $c(M) := c_0 + c_1/M$, finding that this choice yields a good fit to our experimental data (see Appendix~\ref{app:timing_model}, Figure~\ref{fig:e2e-experimental-timing}).
This formula accounts for both a constant overhead present for each micro-batch (e.g. waiting times in inter-accelerator communication) and a processing time proportional to the number of examples in a micro-batch.
Details of the model derivation, micro-batching timing diagrams, and fit of the $c_0, c_1$ parameters to experimental timing data, are presented in Appendix~\ref{app:timing_model}.

\begin{figure}[]
    \centering
    \includegraphics[width=0.7\textwidth]{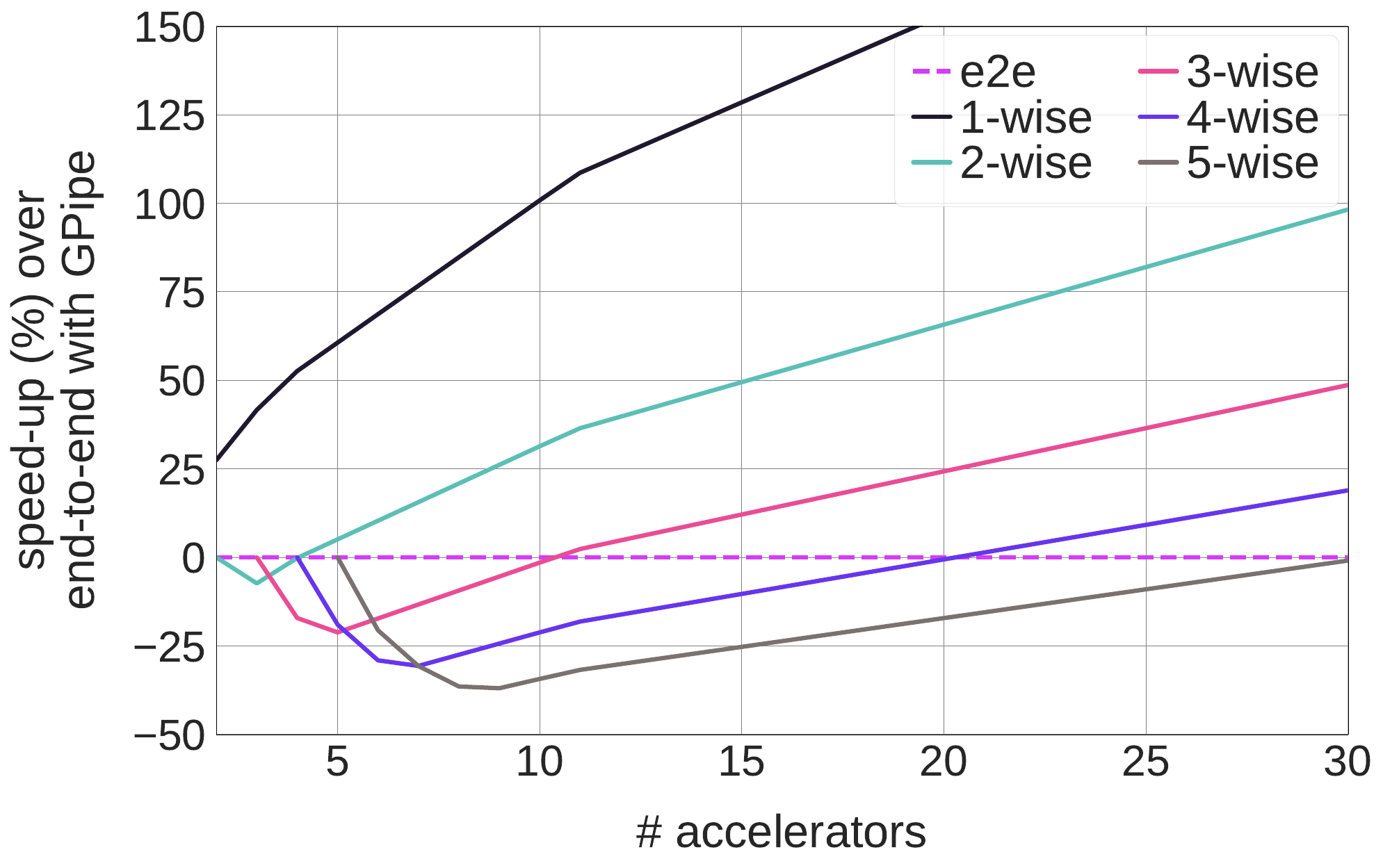}
    \caption{Modelled time per batch speed-up for \nwise{n}, compared to \etoe with GPipe, for varying numbers of accelerators. The optimal number of micro-batches is assumed at each point of the plot (see Appendix~\ref{app:timing_model} Table~\ref{tab:optimal-m-values}). Micro-batch cost ${c(M)=0.025\text{s} + 1.279\text{s}/M}$, tuned to our Transformer experiments, is used. Experimental measurements validating the timing model are presented in Appendix~\ref{app:timing_model}, Figure~\ref{fig:e2e-experimental-timing}.}
    \label{fig:nwise_speedups}
\end{figure}

First, we note that for \nwise{n} with $N < 1 + A/2$ the time per batch is completely independent of the total number of modules of the network.
This demonstrates that the timing benefits of \nwise{n} are most pronounced for big models, distributed across a large number of accelerators.
Next, we compare \nwise{n} with micro-batching to GPipe.
For $c(M)$ as defined above, the benefits of micro-batching are maximised at $M \propto \sqrt A$ for GPipe, while $M=2$ is optimal for $\nwise{n}$ (assuming $N < A/2 + 1$).
For the optimal choice of $M$ in each case, Figure~\ref{fig:nwise_speedups} shows the predicted speed-up of \nwise{n} with micro-batching over GPipe.
For optimal choices of $M$, and the cost function parameters $(c_0=0.025\text{s},c_1=1.279\text{s})$ tuned to results of our Transformer experiments, the model predicts a $50\%$ speed-up of time per batch (compared to end-to-end) for \nwise{2} at $15$ or more accelerators.
This demonstrates that interlocking backpropagation and GPipe are complimentary techniques that are suitable to being combined.
Having demonstrated this, in the rest of the paper we consider the variant of interlocking backpropagation without micro-batching.

\subsection{Information Flow in Interlocking Backpropagation}
Unlike local training, in \nwise{n} interlocking backpropagation the gradient from the loss at the final layer affects all the parameters of the network, it just does so indirectly. In \nwise{2} training, the first module's parameters are updated to optimise the second module's loss. The second module's loss depends on its parameters, which are updated to optimise the third module's loss, and so on, until the final loss. As a result, there is indirect communication from modules at the head of the model to previous modules.
Note that a similar argument about indirect gradient communication in this type of model was made concurrently by \citet{loco}.

Consider a network with $n$ modules trained using \nwise{2}:
\[ f = f_{c_n} \circ \cdots \circ f_{c_1} \]
Each auxiliary network approximates the composition of the modules above it:
\[ h_{c_k} \approx f_{c_n} \circ \cdots \circ f_{c_{k+1}} \]
For training to work effectively, this approximation should be as close as possible, so that the gradient computed from $\mathcal{L}_{c_k}$ encourages $f_{c_{k}}$ to learn a representation which is useful for the modules above. This observation points to a trade-off that arises between modelling quality and compute efficiency, controlled by the size of the auxiliary networks.

In \nwise{1} training, under the particular supervised learning setting we consider in this paper, several factors discourage $h_{c_k}$ from being a good approximation of the remainder of the modules. As $h_{c_k}$ has much lower capacity than the rest of the network in the modules above, it may encourage $f_{c_k}$ to greedily learn a simpler representation which is more amenable to immediately computing the logits, rather than feeding into the subsequent modules. This simpler representation may throw away information which the subsequent modules could use to achieve higher test accuracy.

In \nwise{2} training, we hope that the communication between modules allows upstream modules to learn a representation that is useful for the downstream modules. We argue that this could happen by starting at the head and walking down the model. The penultimate module $f_{c_{n-1}}$ is updated using gradients which have propagated from the true loss at the head of the network and through $f_{c_n}$. Thus, $f_{c_{n-1}}$ is incentivised to learn the most useful representation for $f_{c_{n-1}}$, rather than learning a representation which improves the performance of $h_{c_{n-1}}$. Now we examine the updates of the $f_{c_{n-2}}$. This module is updated with gradients that propagate from $\mathcal{L}_{c_{n-1}}$, and so depend on $h_{c_{n-1}}$. If $h_{c_{n-1}}$ is a close approximation to $f_{c_n}$, then these gradients push $f_{c_{n-2}}$ towards a function which outputs a useful representation for both $f_{c_{n-1}}$ and $f_{c_n}$. As $h_{c_{n-1}}$ and $f_{c_n}$ both have the same inputs and targets we hope that $h_{c_{n-1}}$ should become a close approximation to $f_{c_n}$, as close as possible given the limited capacity of $h_{c_{n-1}}$. Thus, $f_{c_{n-2}}$ should learn a useful representation for $f_{c_{n-1}}$ and $f_{c_n}$. We can continue to extend this reasoning down the model.

\section{Experiments}
\label{sec:experiments}
Here we present results of experiments in both the image and language domains. We compare our \nwise{n} strategies to both extremes of local and end-to-end training.

\subsection{Validation of Approach Using a Small Convolutional Network} \label{CIFAR-10}

\begin{figure}
    \vspace{-1cm}
    \centering
    \includegraphics[width=\textwidth, keepaspectratio]{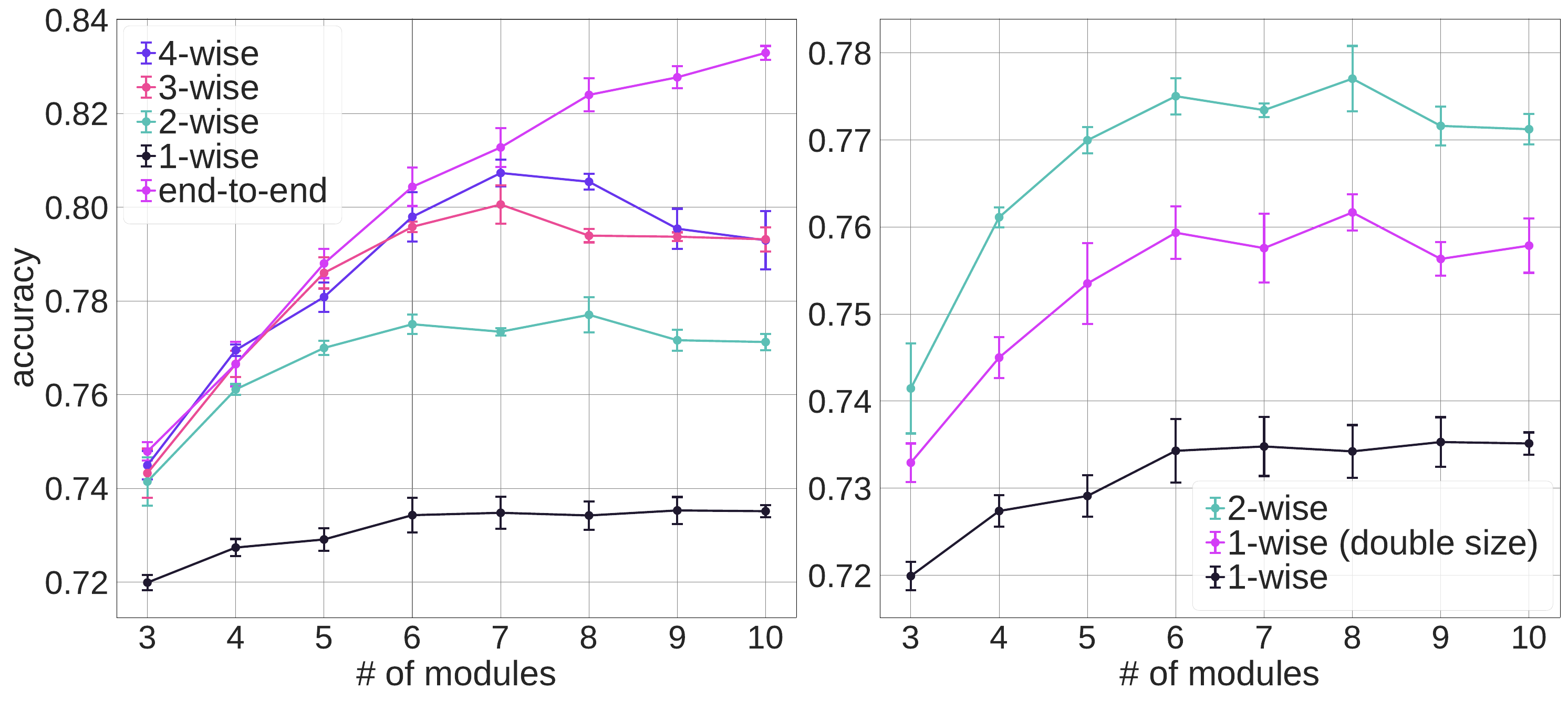}
    \caption{(left) Test accuracy of small convolutional models on CIFAR-10, comparing across depth and training method.
    (right) Comparison demonstrating that simply doubling the size of modules trained locally does not recover the full benefits of \nwise{2} training. In both cases the error bars show one standard deviation over four random seeds.}
    \label{fig:convnet_acc_vs_components}
\end{figure}

We investigate the behaviour of our method when training a small convolutional network on the CIFAR-10 dataset \citep{cifar10}. We consider models of $3$ to $10$ convolutional layers. Each module of the model contains a single convolutional layer and batch norm, with the final two modules also containing max pool layers. The auxiliary networks used for the local losses are comprised of a single linear layer. Appendix~\ref{app:experiment_details} gives full details of the experiment configuration. We train the model using several approaches: \local, \nwise{2}, \nwise{3}, \nwise{4}, and \etoe. \Cref{fig:convnet_acc_vs_components} shows how the different approaches to training perform, as we increase the number of modules in each model. Unsurprisingly, we see that \etoe optimisation results in the best test accuracy, and local optimisation results in the worst. Increasing the $\nwise{n}$ $N$ parameter consistently improves performance, validating our view of interlocking backpropagation as an interpolation between local and global training. For models with 6 or fewer modules, we find that \nwise{3} and \nwise{4} methods achieve comparable accuracy to \etoe training, within standard deviation. A performance gap between \etoe and \nwise{n} for small values of $N$ appears, as we increase the number of modules. This is likely a result of increasing and accumulating approximation errors of the auxiliary networks. However, interlocking backpropagation is shown to remain a viable middle ground between local and global training for all model sizes, as it consistently outperforms local learning, while offering increasing speed-ups over \etoe. While this is only a toy setup, these conclusions are also supported by our examination of ResNets and Transformer networks later in the paper.

In order to understand the interplay between the number and size of the modules in the network and the optimisation strategy, we compare our \nwise{2} training scheme to an alternative approach with similar compute efficiency in \Cref{fig:convnet_acc_vs_components} (right). This approach, labelled `1-wise (double size)', is equivalent to performing \nwise{1} training, but with adjacent pairs of modules merged to give half the number of modules, each of which is twice the size. As merging modules is not possible in practice -- each module would be large enough to fill an entire accelerator -- we could implement this method by grouping modules into blocking pairs. We are interested in a comparison with this method because it is similar to \nwise{2}, except there is no possible indirect communication between modules which are not directly adjacent. For \nwise{2}, we have proposed that the fact that the pairs of modules are overlapping induces indirect communication further down the model than just the adjacent module that the gradients are passed to. \Cref{fig:convnet_acc_vs_components} shows that `1-wise (double size)' performs better than \nwise{1}, but not as well as \nwise{2}, which suggests that there is indirect end-to-end information flow happening in \nwise{2}.

\subsection{Image Domain Results Using ResNet Models}
\label{sec:resnet-experiments}

\begin{table}[]
    \centering
    \begin{tabular}{l c c c c c}
        \toprule
        & & \etoe & \local & \nwise{2} & \nwise{3} \\
        \midrule
        CIFAR-10 & ResNet-32 & 95.20 (0.11) & 94.20 (0.09) & 95.05 (0.09) & \textbf{95.42} (0.06) \\
        CIFAR-100 & ResNet-32 & 76.71 (0.14) & 75.02 (0.09) & \textbf{78.09} (0.13) & 77.84 (0.04) \\
        ImageNet & ResNet-50 & 75.60 & 72.05 & 74.45 & \textbf{76.27} \\
        \bottomrule
    \end{tabular}\\
    \vspace{0.3cm}
    \caption{Accuracy of ResNet-32 and ResNet-50. For CIFAR we give the accuracy on the test set, for ImageNet we give the accuracy on the validation set. One standard error over three seeds is given in brackets for CIFAR. Bold indicates the best performing strategy. The auxiliary networks used are larger than in the toy experiments illustrated in Figure~\ref{fig:convnet_acc_vs_components}, explaining the decreased gap between local and end-to-end learning performance.}
    \label{tab:resnet-results}
    \vspace{-0.6cm}
\end{table}

Having investigated our method in a toy setting, in this section we demonstrate that it continues to lead to improved performance with a more realistic model architecture. In particular, we consider ResNets \citep{he2016deep} on CIFAR-10, CIFAR-100, and ImageNet \citep{deng2009imagenet}. While a ResNet is usually sufficiently small to fit on a single accelerator, these results suggest that our method would work with significantly larger vision models which require multiple accelerators.
In the next section we consider a language model, a Transformer, which is is too large to fit on a single accelerator.

Table \ref{tab:resnet-results} compares the performance of different training schemes for a ResNet-32 and ResNet-50.
In both cases the model is split into four modules by grouping layers with the same feature map size, with the first module containing the input layers, and the last the output layers.
For example, the ResNet-50 we use for the ImageNet experiments is split as follows.
Referring to \citet[Table 1]{he2016deep}, the first module contains the layers labelled `conv1' and `conv2\_x', the second module contains `conv3\_x', the third `conv4\_x', and the forth `conv5\_x' and the output layers.
For the auxiliary classification network we use two convolutional layers with batch norm, global average pooling and a single linear layer. The full experiment configuration is given in Appendix~\ref{app:experiment_details}. These results show that \nwise{n} training substantially closes the performance gap between local and \etoe training. The results also show that \nwise{3} training does not consistently offer better performance than \nwise{2} training which, given the results in Section \ref{CIFAR-10}, is what we would expect for a model with only $4$ modules.

In fact we see that for both CIFAR-10 and CIFAR-100, interlocking backpropagation outperforms \etoe training. This is a surprising finding, as one would expect that a model in which all features were trained to optimise the single global loss would outperform a model in which modules were optimised for local losses. Our results indicate that \nwise{2} training of large scale neural networks could outperform training equivalent models with \etoe backpropagation.
This surprising result may be explained by the success of Inception Nets \citep{inception-nets}, which found that adding auxiliary classification losses into the model improved training performance, leading to state of the art results at the time they were first published.
We have argued that interlocking backpropagation may be able to propagate information from the top level loss to the initial layers, and Inception Nets showed that auxiliary losses improved performance on CIFAR-10. The success of interlocking backpropagation in this setting may be understood as the consequence of these two findings.

\begin{figure}
    \centering
    \includegraphics[width=\textwidth, keepaspectratio]{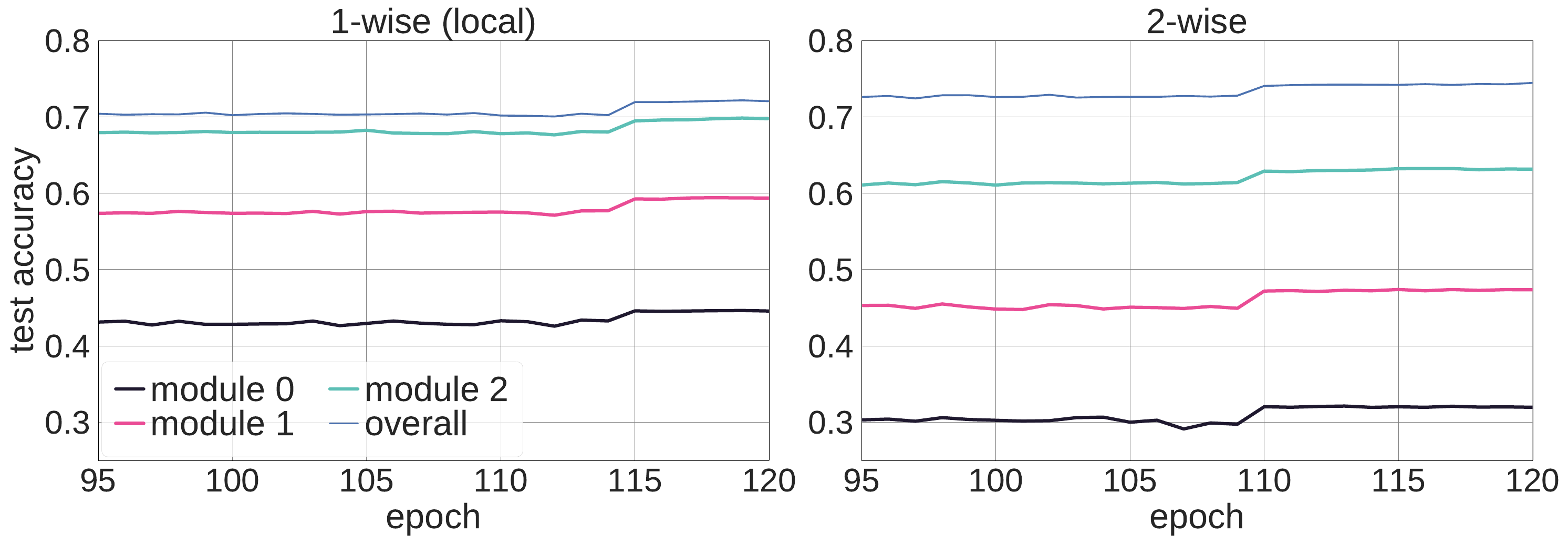}
    \caption{Test accuracy of each auxiliary classification head from a ResNet-50 model trained on ImageNet, trained with \nwise{1} (left) and \nwise{2} (right). \nwise{1} training leads to each module attempting to solve the entire problem on its own; this causes earlier modules to out-perform \nwise{2}, but ultimately, the final performance of \nwise{2}'s more incremental solution strategy is better than \nwise{1}'s.}
    \label{fig:test_accuracy_per_component_resnet}
\end{figure}

We also use these ResNet experiments to further investigate the behaviour of each module in the network.
Figure \ref{fig:test_accuracy_per_component_resnet} shows the test accuracy of the outputs of the auxiliary network of each module in a network trained using \local, compared against a network trained using \nwise{2}.
Examining the accuracy of the model trained using \local, we can see that the accuracy of module 2 is close to the accuracy of the overall model.
This suggests that the model is encountering information loss, as later layers are unable to improve on the representations learned by earlier layers.
The lower modules of the model learn a representation which is suited for the low capacity auxiliary network to map to the logits, throwing away useful information which the subsequent modules could otherwise use to improve accuracy further.
In contrast, if we examine the \nwise{2} performance we notice that the training accuracy increases more gradually with each module, indicating that that this training regime is able to make use of the additional modules to improve performance.

Finally, we compare \nwise{2} to an alternative training method which we refer to as \finetuning, which is another way of interpolating between local and end-to-end learning.
In this method, we first train for $T_1$ epochs using \nwise{1}, then continue training the model using \etoe for $T_2$ epochs.
To compare this to the other training methods, we consider a setting where we have a small time budget within which we wish to achieve the best accuracy possible.
Note that, if a large time budget is available, one can achieve good performance with \finetuning by simply setting $T_1=0$ and training fully with \etoe.
The specific scenario we consider is training a ResNet-32, split into four modules as above, on CIFAR-10 and CIFAR-100.
On CIFAR-10 we consider a time budget corresponding to $80$ epochs of \nwise2 training, and on CIFAR-100 a budget of $100$ epochs of \nwise{2} training.
We convert this into training budgets for the other methods using the table at the bottom of Figure \ref{fig:cost-comparison}.
For example, for CIFAR-100 we run $50$ epochs of \etoe training and $200$ epochs of \nwise1 training.
In the case of \finetuning, the total number of epochs is $T_1 + T_2$, where $T_1$ is a hyperparameter and $T_2 = (2 \cdot (\text{num \nwise2 epochs}) - T_1)/4$.
In addition to $T_1$, another important hyperparameter for \finetuning is the learning rate used at the start of the \etoe phase, which we denote by $\eta_2$.
We select both $T_1$ and $\eta_2$ using a grid search, the results of which are given in Table \ref{tab:fine-tuning-grid-search} in Appendix \ref{app:additional_results}.
Table~\ref{tab:fine-tuning} compares the test accuracy of \finetuning to \nwise1, \nwise2, and \etoe.
It reveals that, in these time constrained scenarios, both \nwise2 and \finetuning are able to outperform \nwise1 and \etoe training.
Taking account of the standard error, \nwise2 and \finetuning perform similarly, though \finetuning has a higher mean accuracy on the less complex task (CIFAR-10) and smaller time budget, and \nwise2 a higher mean accuracy on the more complex task (CIFAR-100) and slightly larger time budget.
However, achieving this performance with \finetuning requires a large grid search to choose $T_1$ and $\eta_2$, which makes it difficult to apply to large models that require a large amount of compute to train.
In particular, we were not able to evaluate \finetuning on the language models featured later in this paper due to the computational cost.
Full details of the \finetuning algorithm, and other experiment configuration, are given in Appendix \ref{app:experiment_details}.

\begin{table}
    \centering
    \begin{tabular}{l c c c c}
        \toprule
                  & \etoe        & \nwise2               & \local       & \finetuning           \\
        \midrule
        CIFAR-10  & 92.86 (0.15) & 94.06 (0.12)          & 93.98 (0.07) & \textbf{94.21 (0.07)} \\
        CIFAR-100 & 73.98 (0.13) & \textbf{76.85 (0.14)} & 74.33 (0.09) & 76.40 (0.16)          \\
        \bottomrule
    \end{tabular}
    \caption{
        Test accuracy when training under a fixed time budget, including results using \finetuning.
        In brackets we give the standard error over $4$ random seeds, and bold indicates the highest mean.
    }
    \label{tab:fine-tuning}
\end{table}

\subsection{Language Domain Results Using Transformer Models}
\label{subsection:transformer-results}

\begin{figure}[t]
    \centering
    \includegraphics[width=0.49\textwidth]{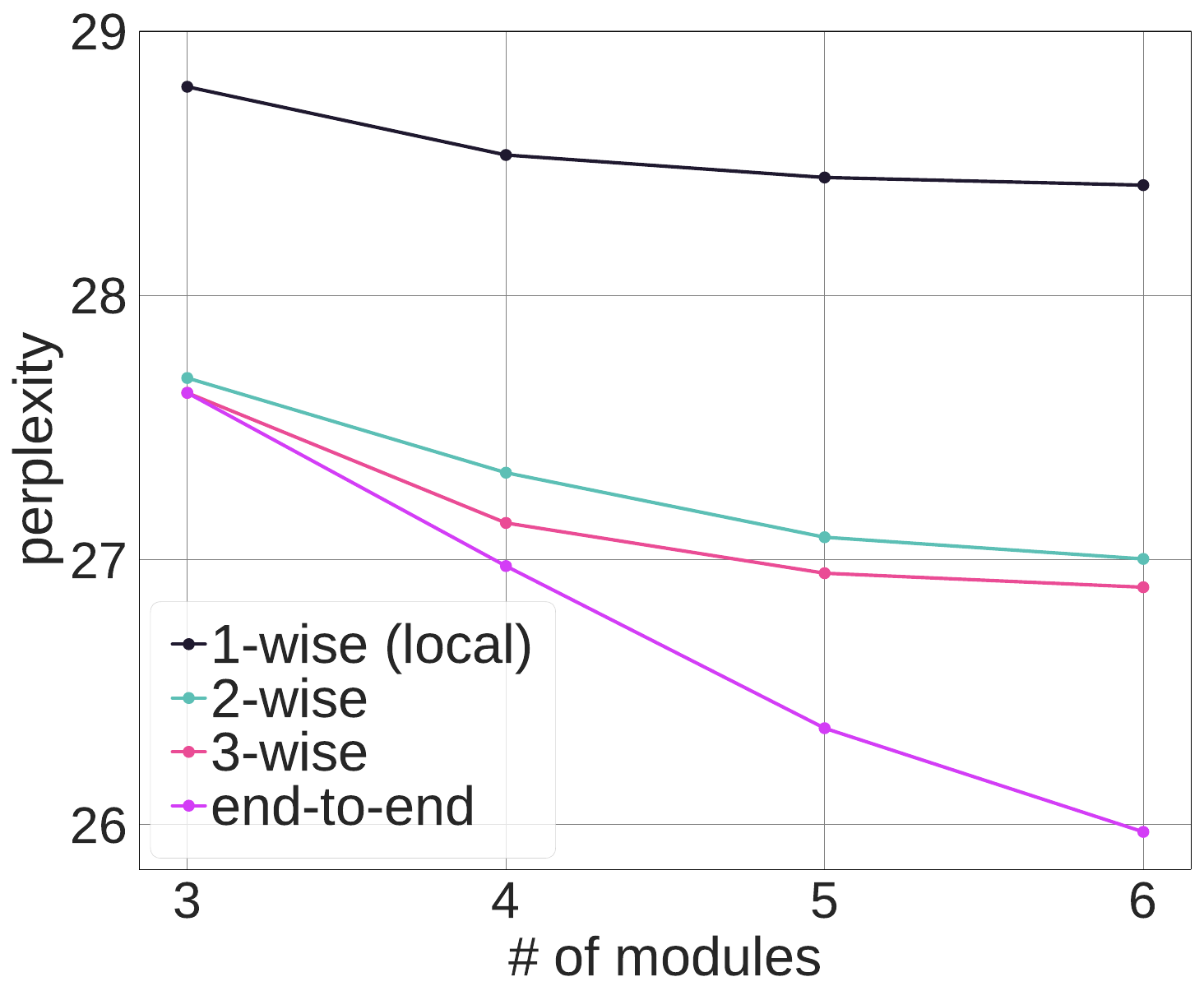}    \includegraphics[width=0.49\textwidth, keepaspectratio]{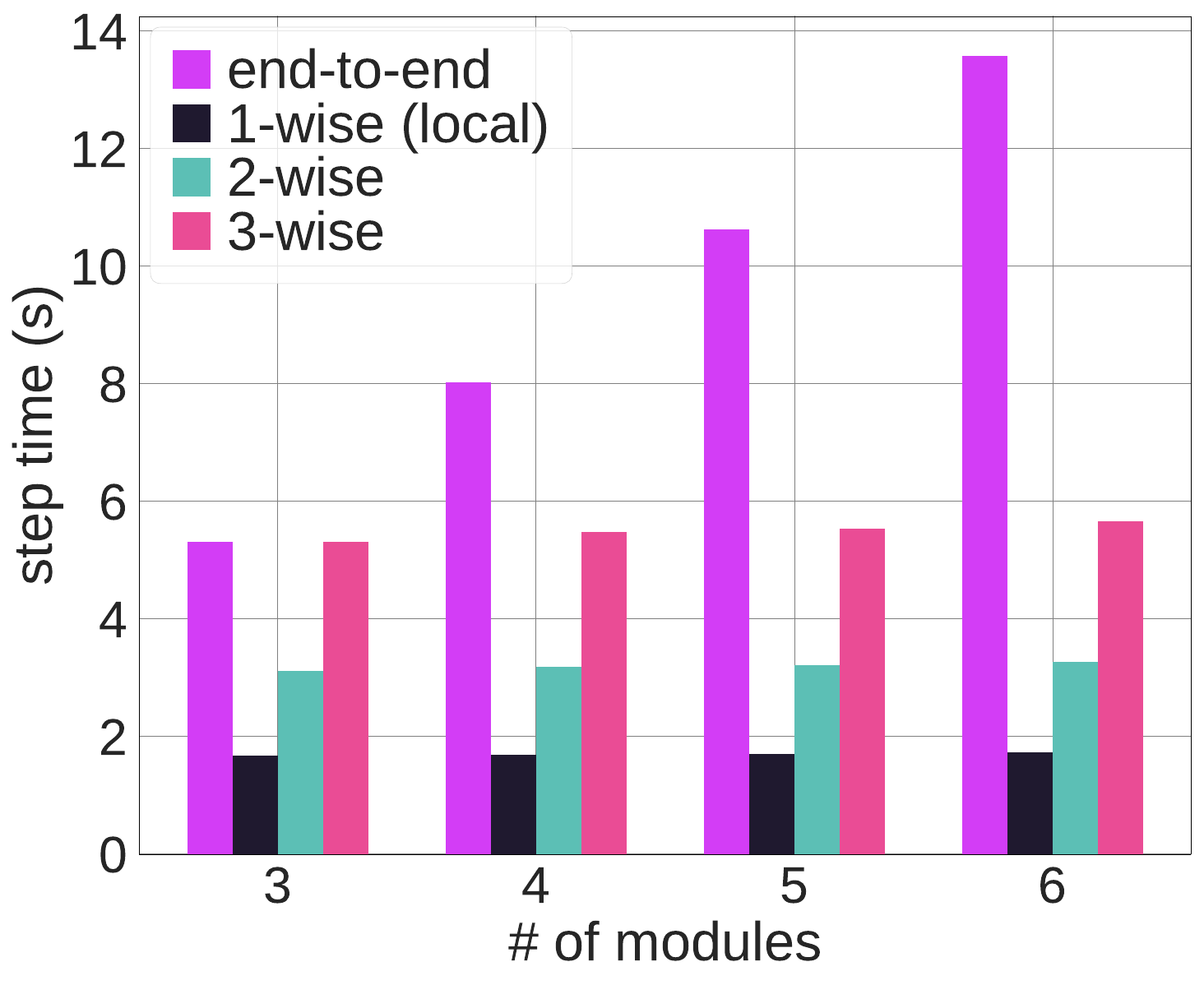}
    \caption{(left) Comparison of test perplexity of Transformer models across depth and training method, for a single simulation in each configuration.
    Models are run for a \emph{fixed number of steps}.
    \nwise{2} recovers a large amount of model performance lost in local training.
    (right) Comparison of training step time (in wall-clock seconds) of Transformer-based models across depth and training method.
    }
    \label{fig:train_accuracy_per_component}
\end{figure}

\begin{figure}[t]
    \centering
    \includegraphics[width=0.49\textwidth, keepaspectratio]{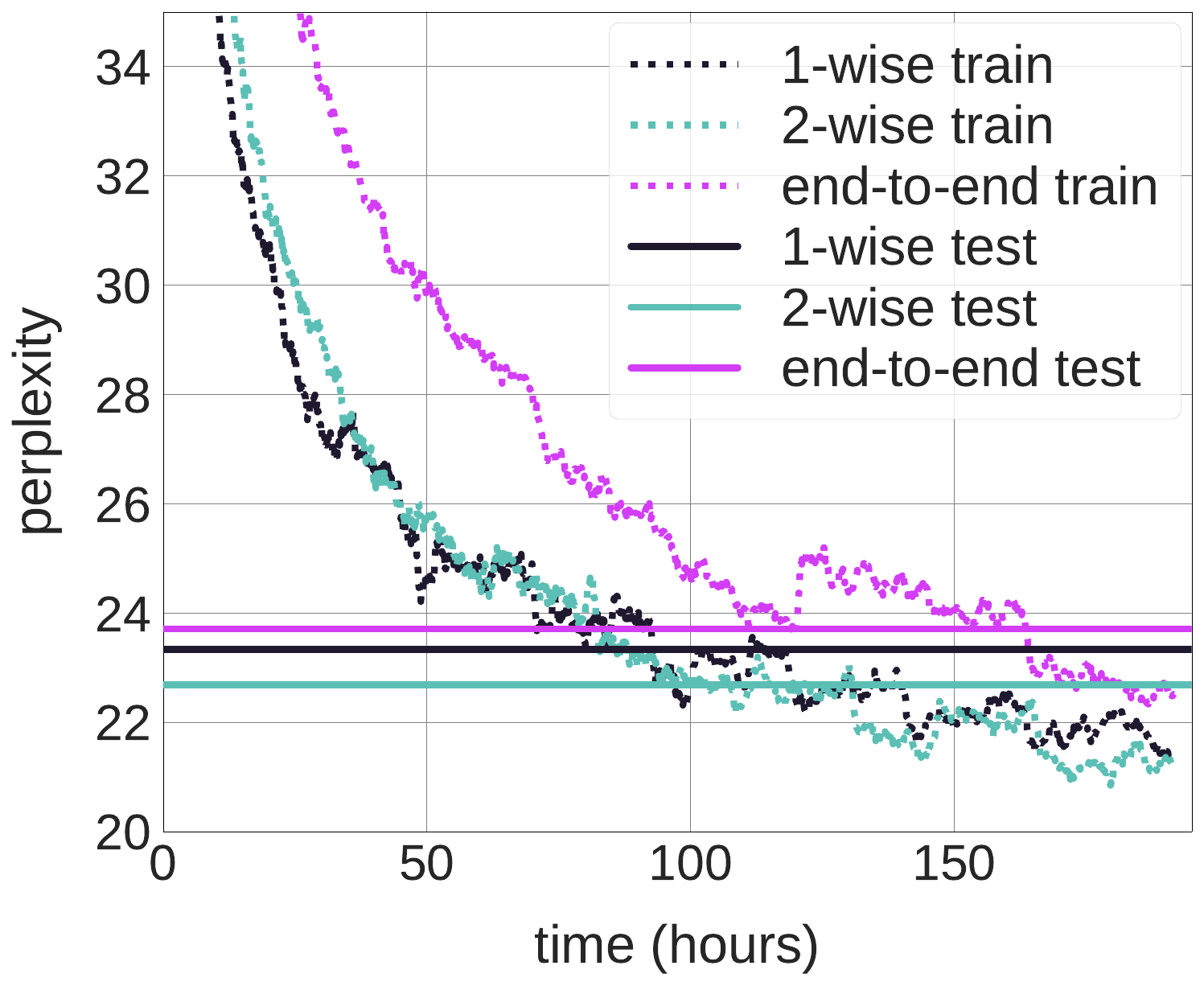}
    \includegraphics[width=0.49\textwidth]{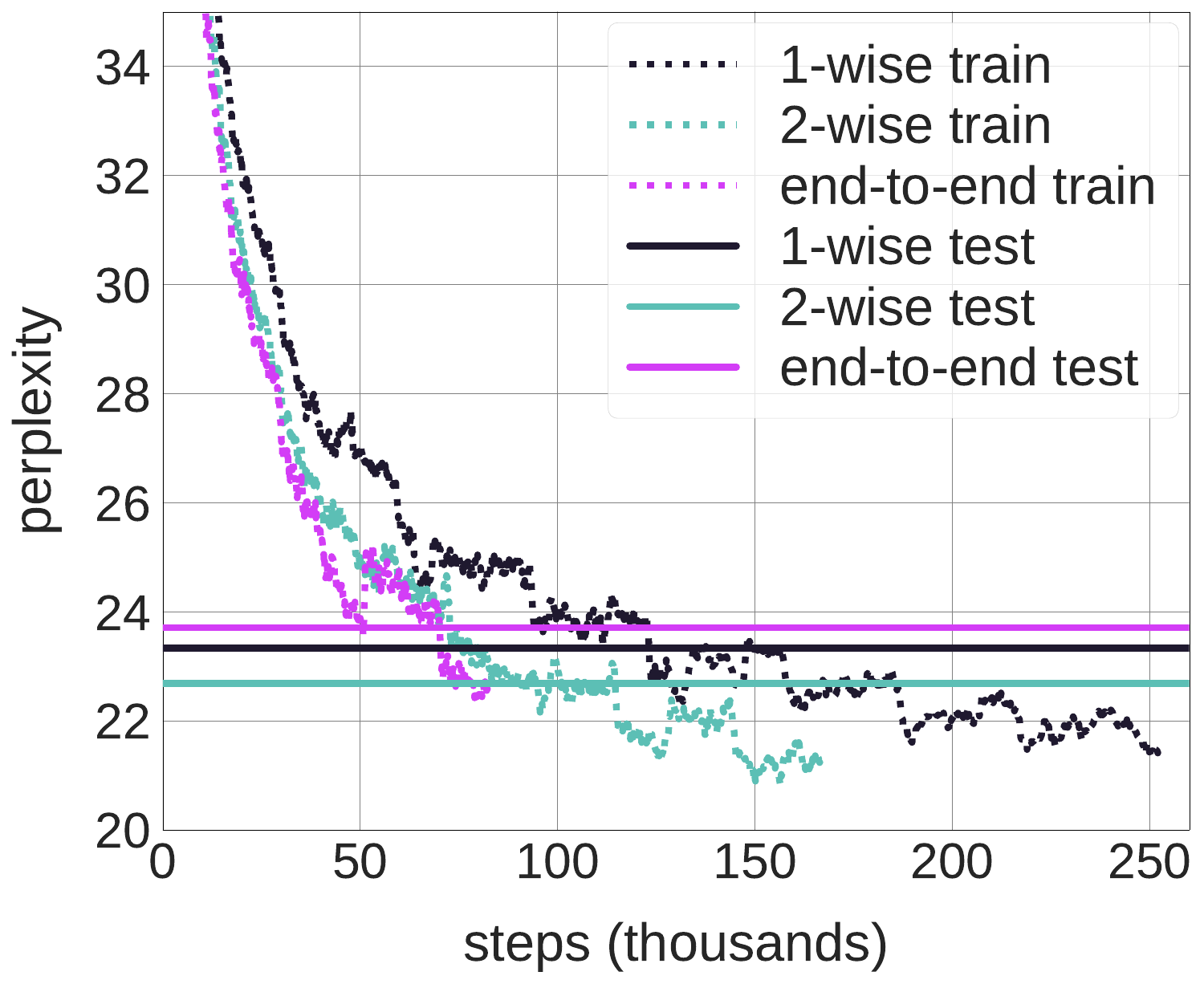}
    \caption{While Figure~\ref{fig:train_accuracy_per_component}-(left) may suggest that end-to-end training always out-performs local training, it is important to keep in mind that this is still in the `fixed steps' perspective; i.e we fix the number of steps each model is allowed to take and ignore the fact that running times may differ wildly.
    This figure shows the training (dashed lines) and test (solid lines) perplexities (measured at the end of training), of Transformer models comprised of four modules and trained for eight days.
    The results are reported for a single simulation in each configuration.
    It demonstrates the importance of considering the `fixed time' perspective: (left) depicts the training curves of \nwise{1}, \nwise{2}, and \etoe in a `per time' perspective; (right) depicts the same training curves in a `per step' perspective.
    The difference between these two perspectives is quite extreme -- from the `per step' perspective, end-to-end training is best at any given point; however, when a `per time' perspective is considered, \nwise{1} and \nwise{2} are best at any given point.
    Given a fixed time constraint, the logical decision to obtain the best possible model is to opt for an \nwise{n} strategy.}
    \label{fig:time-versus-steps}
    \vspace{-0.5cm}
\end{figure}

Finally we investigated the performance of our method for training Transformer based models on language modelling tasks. The architecture we used largely follows the decoder only Transformer described in OpenAI's GPT-2 model \citep{radford2019language}, with the addition of the auxiliary classification networks used for calculating the local losses.
See Appendix \ref{section:transformer_exp} for details.
In these experiments each module is made out of 6 Transformer blocks. We run experiments with networks comprised of 3 to 6 modules. Each module was trained on a v3-8 TPU. We trained and evaluated the models with the One Billion Word Benchmark for Language Modelling \citep{lm1b}. Each Transformer block module has a dimensionality of 1024. We train with a max sequence length of 128, and a batch size of 1024. For the experiments in Figure~\ref{fig:train_accuracy_per_component}, we train for one epoch with the Adam optimiser; for the experiments in Figure~\ref{fig:time-versus-steps} we train for 192 hours (eight days).
We measure the performance of the models using perplexity, for which a lower value indicates better performance.
Figure~\ref{fig:train_accuracy_per_component} shows the test set perplexity of models trained with \nwise{1}, \nwise{2} interlocking, \nwise{3} interlocking and, \etoe backpropagation.
We can see that, unsurprisingly, \etoe greatly outperforms \nwise{1} training and the gap between them widens as we increase the size of the model. Interlocking backpropagation is able to make up much of the gap between \nwise{1} and \etoe, but unlike our observations with CIFAR-10 and CIFAR-100, there is still a test perplexity gap between interlocking and \etoe backpropagation for a fixed number of training steps.

In this real world setting we are able to substantially decrease the training time of these large models. The time per step for models of this size varies considerably based on the optimisation strategy used. Figure \ref{fig:train_accuracy_per_component} visualises the test set perplexity and the train step time for models of various sizes trained with \etoe, \nwise{1}, and \nwise{2} interlocking backpropagation. \nwise{2} interlocking backpropagation requires less than half the training step time of standard \etoe training, to achieve similar test perplexity for models with 4 modules.

Figure \ref{fig:time-versus-steps} demonstrates the importance of considering a `fixed time' perspective of training. Instead of fixing the number of optimiser steps, we fix the total elapsed training time to a set number of hours. The result is that methods, which take gradient steps quicker, see many more weight updates relative to slower methods. When running for a fixed amount of time, the performance of interlocking backpropagation improves dramatically relative to end-to-end methods.

\section{Conclusion}
We have explored a variety of optimisation strategies, for large scale, distributed neural networks, that strike a middle-ground between local and end-to-end training.
These strategies, referred to as \nwise{n} training, introduce intermediate auxiliary classification heads and losses into the network and train with interlocking backpropagation, which restricts the gradient flow between accelerators.
We have shown that \nwise{2} interlocking backpropagation significantly reduces training time of large scale distributed neural networks, and recovers much of the test accuracy that is lost in local training.
\nwise{2} training even outperforms \etoe on ResNets trained on CIFAR-10 and CIFAR-100, though it does not outperform baselines in language modelling tasks with Transformers.
We have provided evidence that varying the $N$ parameter of \nwise{n} can be seen as interpolating between \local and \etoe training, in terms of compute efficiency and modelling performance.
Interlocking backpropagation has been shown to be a practical approach to training large scale distributed neural networks.

\newpage
\section*{Acknowledgments}
Oscar Key acknowledges funding from the Engineering and Physical Sciences Research Council (EPSRC), grant number EP/S021566/1.

\bibliography{references}

\begin{appendices}

\section*{Supplementary Material}

\section{Additional Results}
\label{app:additional_results}

\begin{table}[h]
    \centering
    \begin{tabular}{lllll}
        \toprule
        & \multicolumn{2}{l}{ImageNet} & \multicolumn{2}{l}{CIFAR-10} \\
        & \nwise{2}        & \local       & \nwise{2}         & \local       \\
        \midrule
        top 1 & \textbf{74.45}        & \textbf{72.05}       & 95.12        & 94.78       \\
        top 2 & 73.46       & 71.70       & 95.10         & \textbf{94.89}       \\
        top 3 & 72.46       & 71.34      & 95.18        & 94.76       \\
        top 4 & 71.75       & 70.92      & \textbf{95.26}        & 94.82       \\
        \bottomrule
    \end{tabular}
    \vspace{0.3cm}
    \caption{Accuracy when ensembling the predictions made by the top $n$ modules in the model, when using \nwise{2} interlocking backprop model and a \nwise{1} local training. On CIFAR-10 we report accuracy on the test set, on ImageNet we report validation on validation set. We present this as a negative result, demonstrating that ensembling the classification heads does not lead to noticeably better performance. In the case of ImageNet, ensembling actually hurt performance.}
    \label{tab:ensembled-predictions}
\end{table}

\begin{table}[h]
    \centering
    \begin{tabular}{l c c c}
        \toprule
        \# Modules & 3 & 4 & 5 \\
        \midrule
        Perplexity & 82.682 & 100.183 & 110.278 \\
        \bottomrule
    \end{tabular}\\
    \vspace{0.3cm}
    \caption{The extremely poor task performance of Hogwild optimisation due to gradient staleness (see Fig.~\ref{fig:hogwild}). The model is the same Transformer used in Figure \ref{fig:train_accuracy_per_component} with the same optimisation parameters and number of training epochs. Increasing the number of modules used causes the average staleness for modules to increase, leading to extremely unstable training dynamics that makes optimisation difficult. Despite Hogwild improving training speeds to a similar effect as local learning, it is ultimately too poor at model optimisation to be considered a viable alternative to end-to-end learning.}
    \label{tab:transformer-hogwild}
\end{table}

\begin{table}[h]
    \centering
    \begin{tabular}{rlcc}
        \toprule
        $T_1$ & $\eta_2$ &
        \multicolumn{2}{c}{val. accuracy}                                \\
              &          & CIFAR-10              & CIFAR-100             \\
        \midrule                                                         \\
        60.0  & 0.1      & 93.78 (0.10)          & -                     \\
              & 0.01     & 93.31 (0.09)          & -                     \\
        80.0  & 0.1      & \textbf{93.86 (0.13)} & \textbf{77.00 (0.21)} \\
              & 0.01     & 93.40 (0.12)          & 75.66 (0.19)          \\
        100.0 & 0.1      & 93.70 (0.10)          & 76.71 (0.27)          \\
              & 0.01     & 93.56 (0.11)          & 75.46 (0.16)          \\
        120.0 & 0.1      & 93.53 (0.13)          & 76.78 (0.22)          \\
              & 0.01     & 93.45 (0.06)          & 75.42 (0.17)          \\
              & 0.001    & 93.46 (0.14)          & -                     \\
        140.0 & 0.1      & -                     & 76.10 (0.13)          \\
              & 0.01     & 93.26 (0.10)          & 75.24 (0.07)          \\
              & 0.001    & 93.34 (0.06)          & -                     \\
        160.0 & 0.1      & -                     & 76.16 (0.17)          \\
              & 0.01     & -                     & 75.32 (0.17)          \\
              & 0.001    & -                     & 74.57 (0.21)          \\
        180.0 & 0.01     & -                     & 74.72 (0.18)          \\
              & 0.001    & -                     & 74.50 (0.02)          \\
        \bottomrule
    \end{tabular}
    \caption{
        Results of a grid search to choose the hyperparameters for the \finetuning results in Table \ref{tab:fine-tuning}.
        We search over $T_1$ -- the number of epochs of \nwise1 training -- and $\eta_2$ -- the learning rate at the start of the \etoe phase of training.
        Bold indicates the optimal configuration, and we give the standard error over four random seeds in brackets.
    }
    \label{tab:fine-tuning-grid-search}
\end{table}

\section{Experiment Details}
\label{app:experiment_details}

\subsection{\Cref{fig:convnet_acc_vs_components} (small convolutional network experiments)}

The configuration of an $n$-module main network is as follows, where each box represents a module:

\begin{center}
\begin{tabular}{|c|}
	\specialrule{1.5pt}{0pt}{0pt}
	\hicell logits \\
	\specialrule{1.5pt}{0pt}{0pt}
	linear \\
	maxpool2d kernel=2x2 stride=1 \\
	conv2d kernel=3x3 filters=64 padding=1 stride=1 (batch norm, ReLU) \\ \hline
	maxpool2d kernel=2x2 stride=1 \\
	conv2d kernel=3x3 filters=64 padding=1 stride=1 (batch norm, ReLU) \\ \hline
	conv2d kernel=3x3 filters=32 padding=1 stride=1 (batch norm, ReLU) \\
	(this module repeats $n-3$ times) \\ \hline
	\ldots \\ \hline
	conv2d kernel=3x3 filters=32 padding=1 stride=1 (batch norm, ReLU) \\
	\specialrule{1.5pt}{0pt}{0pt}
	\hicell input image ($32 \times 32$) \\
	\specialrule{1.5pt}{0pt}{0pt}
\end{tabular}
\end{center}

\noindent
For the \local and \nwise{n} configurations, the auxiliary network is a single linear layer.

\noindent
We use the Adam optimizer with a learning rate of $0.0001$. We train for $100$ epochs. We do not use learning rate decay, weight decay, or data augmentation. We train on the entire training set, and report results on the test set. We normalize the inputs based on the mean and standard deviation of the training set.

\subsection{\Cref{tab:resnet-results}, \Cref{tab:fine-tuning}, and \Cref{fig:convnet_acc_vs_components} (ResNet experiments)}
We use the CIFAR-10, CIFAR-100 and ImageNet ResNet architectures as given by \citet{he2016deep}.
For the \local and \nwise{n} configurations, the auxiliary network is:

\begin{center}
\begin{tabular}{|c|}
	\specialrule{1.5pt}{0pt}{0pt}
	\hicell approximated logits \\
	\specialrule{1.5pt}{0pt}{0pt}
	linear \\
	global average pool \\
	conv2d kernel=3x3 filters=64 padding=1 stride=1 (batch norm, ReLU) \\
	conv2d kernel=3x3 filters=128 padding=1 stride=1 (batch norm, ReLU) \\
	\specialrule{1.5pt}{0pt}{0pt}
	\hicell main network output \\
	\specialrule{1.5pt}{0pt}{0pt}
\end{tabular} \\
\end{center}

\noindent
The training configuration shared between experiments in this section is as follows:
\begin{center}
\begin{tabular}{r c c}
    \toprule
     & \textbf{CIFAR} & \textbf{ImageNet} \\
    \midrule
    optimizer & \multicolumn{2}{c}{SGD} \\
    initial learning rate & \multicolumn{2}{c}{$0.1$} \\
    weight decay & \multicolumn{2}{c}{$0.0002$} \\
    momentum & \multicolumn{2}{c}{$0.9$} \\
    \addlinespace[0.8em]
    data augmentation & \parbox{5cm}{random horizontal flip, $32 \times 32$ crop with padding 4} & \parbox{5cm}{random resized crop of $224 \times 224$, random horizontal flip} \\
    \addlinespace[0.8em]
    test data preprocessing & - & \parbox{5cm}{resize so that shortest side has length $224$, take $224 \times 224$ center crop} \\
    \addlinespace[0.8em]
    batch size & 128 & 256 \\
    \bottomrule
\end{tabular}
\end{center}
We normalize the inputs based on the mean and standard deviation of the training set.

\subsubsection{\Cref{tab:resnet-results} and \Cref{fig:convnet_acc_vs_components}}
In addition to the training configuration above, for these experiments we use the following configuration:
\begin{center}
\begin{tabular}{r c c}
    \toprule
     & \textbf{CIFAR} & \textbf{ImageNet} \\
    \midrule
    learning rate schedule & \parbox{5cm}{divide by $10$ at epochs $[91, 136, 182]$} & \parbox{5cm}{divide by $10$ if the validation loss does not improve for $10$ epochs} \\
    total epochs & 200 & 120 \\
    \bottomrule
\end{tabular}
\end{center}

\noindent
For CIFAR we train on the entire training set, and report results on the test set. For ImageNet we train on the test set, and report results on the validation set.

\subsubsection{\Cref{tab:fine-tuning} (Fine-tuning experiments)}
The \finetuning algorithm is
\begin{enumerate}
    \item Train for $T_1$ epochs using \nwise1.
    \item Discard the auxiliary networks, set the learning rate to $\eta_2$, and reset any optimizer state.
    \item Train for $T_2$ epochs using \etoe.
\end{enumerate}

As discussed in the main paper body, for CIFAR-10 we train for the equivalent of $80$ \nwise2 epochs, and for CIFAR-100 the equivalent of $100$ \nwise2 epochs.
The number of epochs for each method is then as follows:
\begin{center}
    \begin{tabular}{lcc}
        \toprule
        & CIFAR-10 & CIFAR-100 \\
        \midrule
        \etoe & 40 & 50 \\
        \nwise2 & 80 & 100 \\
        \local & 160 & 200 \\
        \finetuning & \multicolumn{2}{c}{$T_1 + T_2$} \\
        \bottomrule
    \end{tabular}
\end{center}
For \finetuning, $T_1$ is a hyperparameter, with $T_2 = (2 \cdot 80 - T_1)/4$ for CIFAR-10 and $T_2 = (2 \cdot 100 - T_1)/4$ for CIFAR-100.

One difficulty with obtaining the results in \Cref{tab:fine-tuning} is selecting the learning rate schedule for each configuration, because the large number of configurations means it is too expensive to optimize it in each case.
Thus we use a heuristic approach to set the schedule.
For \etoe, \nwise2 and \nwise1, we start with a learning rate of $0.1$ and reduce by a factor of $10$ at epochs $0.7n$, $0.85n$ and $0.95n$, where $n$ is the total number of epochs.
For \finetuning we follow the same learning rate schedule as \nwise1 for the first phase.
For the second phase we start at a learning rate of $\eta_2$, and reduce the learning rate by a factor of $10$ at approximately epochs $T_1 + 0.7 T_2$, $T_1 + 0.85 T_2$ and $T_1 + 0.95 T_3$.
The specific epochs at which we reduce the learning rate are given in \Cref{tab:fine-tuning-lr-drops}.
\begin{table}[h]
    \centering
    \begin{tabular}{lrrl}
        \toprule
                  & $T_1$ & $\eta_2$ & lr reduced at $T_1 + $ \\
        \midrule
        CIFAR-10  & 60    & 0.010    & 19, 23                 \\
                  &       & 0.100    & 18, 21, 24             \\
                  & 80    & 0.010    & 14, 18                 \\
                  &       & 0.100    & 12, 16, 18             \\
                  & 100   & 0.010    & 10, 13                 \\
                  &       & 0.100    & 9, 12, 14              \\
                  & 120   & 0.001    & 8                      \\
                  &       & 0.010    & 6, 8                   \\
                  &       & 0.100    & 5, 7, 9                \\
                  & 140   & 0.001    & 4                      \\
                  &       & 0.010    & 3                      \\
        CIFAR-100 & 80    & 0.010    & 22, 27                 \\
                  &       & 0.100    & 20, 26, 28             \\
                  & 100   & 0.010    & 19, 23                 \\
                  &       & 0.100    & 18, 21, 24             \\
                  & 120   & 0.010    & 14, 18                 \\
                  &       & 0.100    & 12, 16, 18             \\
                  & 140   & 0.010    & 10, 13                 \\
                  &       & 0.100    & 9, 12, 14              \\
                  & 160   & 0.001    & 8                      \\
                  &       & 0.010    & 6, 8                   \\
                  &       & 0.100    & 5, 7, 9                \\
                  & 180   & 0.001    & 4                      \\
                  &       & 0.010    & 3, 4                   \\
        \bottomrule
    \end{tabular}
    \caption{
        Epochs at which we reduce the learning rate by a factor of $10$ during the \etoe phase of \finetuning.
    }
    \label{tab:fine-tuning-lr-drops}
\end{table}

For these experiments we withheld $10$\% of the training set to create a validation set, which we used to perform the grid search.

\subsection{Transformer Experiments}
\label{section:transformer_exp}

The main network's Transformer blocks have $4$ attention heads, an embedding dimension of $1024$, and sequence length of $128$. Each module is composed of $6$ such Transformer blocks, amounting to $152\text{M}$ parameters per module, including the auxiliary networks. The first and last modules additionally contain input and output projection layers, respectively.

The architecture of the auxiliary network for each module in the Transformer experiments is:
\begin{center}
\begin{tabular}{|c|}
	\specialrule{1.5pt}{0pt}{0pt}
	\hicell approximated logits \\
	\specialrule{1.5pt}{0pt}{0pt}
	embedding matrix \\
	Transformer block(attention heads=4, embedding dim=1024, max sequence length=128) x 2 \\
	\specialrule{1.5pt}{0pt}{0pt}
	\hicell main network output \\
	\specialrule{1.5pt}{0pt}{0pt}
\end{tabular}
\end{center}

\noindent
The training configuration is as follows:
\begin{center}
    \begin{tabular}{l c c c}
        \toprule
        optimizer & $\operatorname{Adam}(\beta_1 = 0.9, \beta_2 = 0.98, \epsilon = 10^{-9})$ \\
        learning rate & $1024^{-0.5} \times \min( \textrm{step}^{-0.5}, \textrm{step} \times \textrm{warmup\_steps}^{-1.5})$ \\
        lr warmup steps & 4000 \\
        total epochs & 1\\
        \bottomrule
    \end{tabular}
\end{center}

For \nwise{n} Transformer experiments, we additionally use the gradient signal from the auxiliary network local to each module to update the parameters.
Specifically, when updating the parameters of the $k^\text{th}$ module, we take the mean of the gradients propagated from both the auxiliary network in the $k^\text{th}$ module, and from the auxiliary network of the $(k+N-1)^\text{th}$ module.
We found that this improved modelling performance, without affecting the step time.


\section{The Timing Model}
\label{app:timing_model}

\subsection{Deriving the time per batch formula}
\label{ssec:time_per_batch_formula_derivation}

Recall the formula for time per batch, given the number of accelerators $A$, number of micro-batches $M$, the \nwise{n} parameter $N$, and the cost of processing a single micro-batch $c(M)$:

\begin{equation*}
T(A,M,N) = \begin{cases}
    (2 + A-N) \cdot Mc(M) + 2(2N-A-1)\cdot c(M) & \text{(for } M > 2, A < 2N-1 \text{)} \\
    (N + 1) \cdot Mc(M) & \text{(for } M > 2, A \ge 2N-1 \text{)} \\
    (N + 1) \cdot Mc(M) & \text{(for } M = 2 \text{)} \\
    2N \cdot Mc(M) & \text{(for } M = 1 \text{)}
    \end{cases}
\end{equation*}

\begin{figure}
    \centering
    \includegraphics[width=\textwidth]{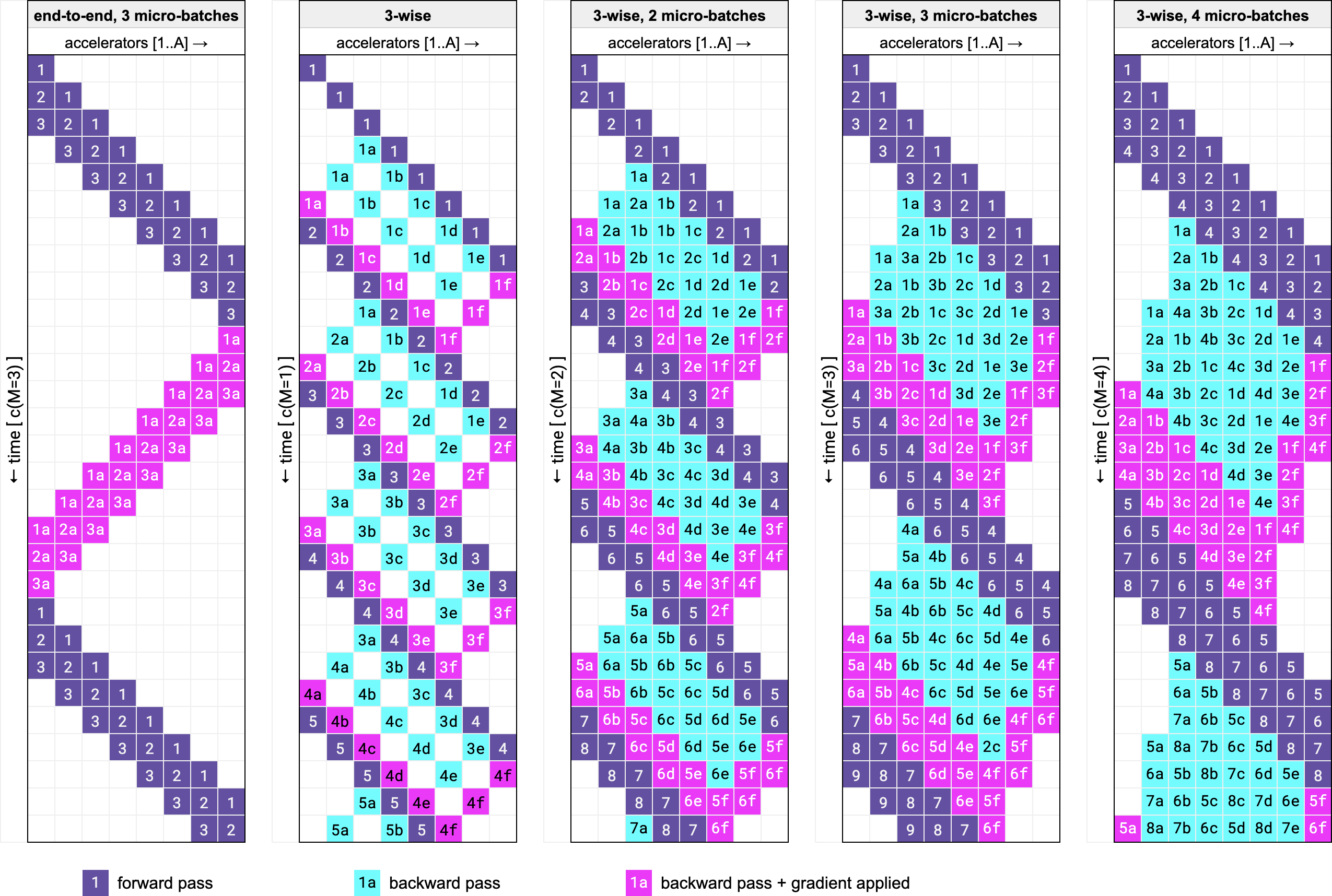}
    \caption[]{End-to-end and \nwise{3} training with GPipe micro-batching. The number in each square indexes the micro-batches, the letter indexes different backward passes of a given micro-batch. The vertical time axis in each diagram is scaled by the micro-batch cost $c(M)$. The larger the $M$, the shorter the duration of a single grid cell in seconds.}
    \label{fig:microbatching-grid-diagrams}
\end{figure}

In reference to the timing grid diagrams, such as pictured in Figure~\ref{fig:microbatching-grid-diagrams}, the quantity $T(A,M,N)/c(M)$ is the smallest period of the timing diagram, expressed in number of grid cells.

\textbf{Case 1: $(M=1)$.} Before the first accelerator can start processing the next mini-batch, the previous micro-batch must undergo $N$ forward passes, and then $N$ backward passes, where the last one triggers the gradient update on the first accelerator.

\textbf{Cases 2 \& 3: $(M=2)$ or $(M>2, A \ge 2N-1)$.} The middlemost accelerator needs to process the forward pass and $N$ different backward passes for each micro-batch, adding up to $(N+1)M$ time slots.

\textbf{Case 4: $(M>2, A < 2N-1)$.} There are $A-(N-1) < N$ accelerators that need to compute their local loss. Consider any of those $A-(N-1)$ accelerators that's not the final one; it needs to compute exactly one more backward pass than its successor, and that makes it have to delay the computation of the first backward pass it receives from its successor by $M-2$ time slots.
Decreasing $A$ by $1$ eliminates one of those delays and thus decreases the total time per batch by $M-2$ time slots. For $M>2, A <2N-1$, and using the formula for (end-to-end) $T(N,M,N)$ easily derivable from the diagrams, we get:

\begin{align*}
T(A,M,N) &= (A-N)(M-2)c(M) + T(N,M,N)
\\&=(A-N)(M-2)c(M) + 2(M+N-1)\cdot c(M)
\\&=(2+A-N) \cdot Mc(M) + 2(2N-A-1)\cdot c(M)
\end{align*}

\subsection{Comparison of timing model with experimental data}

\begin{figure}
    \centering
    \includegraphics[width=.49\textwidth]{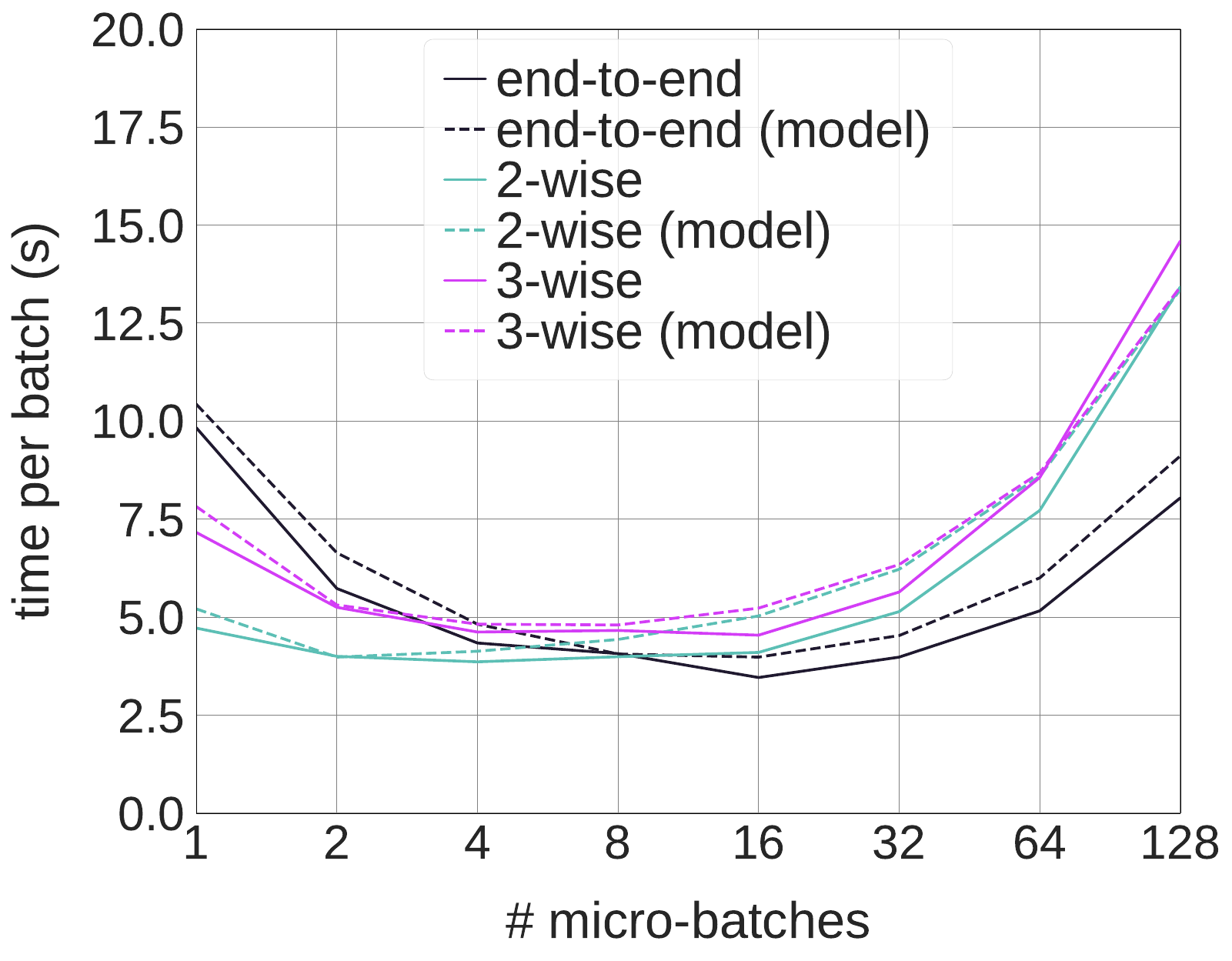}
    \includegraphics[width=.49\textwidth]{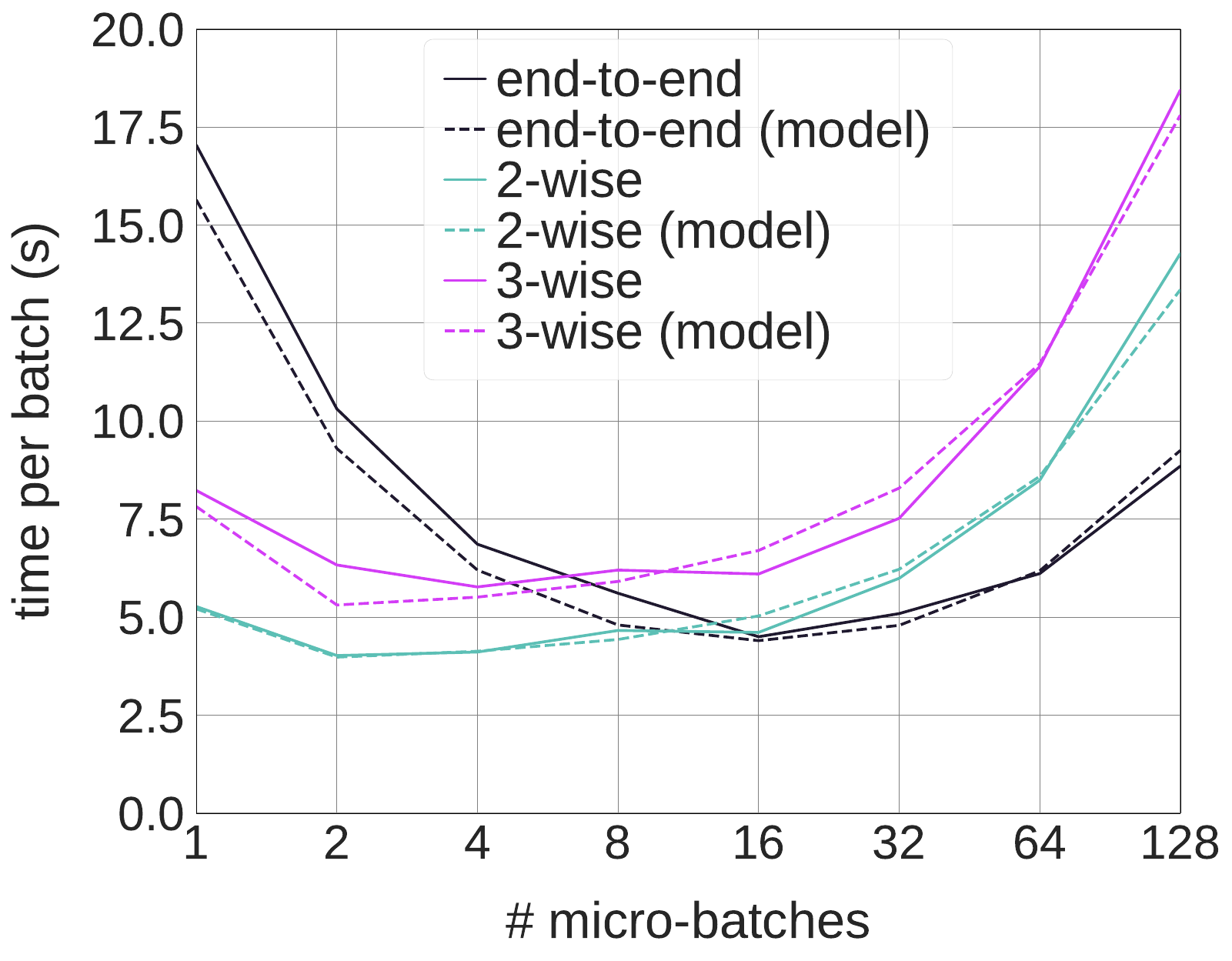}
    \caption[]{Comparison of experimental and modelled training time per batch, across different micro-batching granularities. (left) $A=4$ accelerators, (right) $A=6$. The micro-batch cost parameters were fit to ${c_0=0.025\text{s}, c_1=1.279\text{s}}$.}
    \label{fig:e2e-experimental-timing}
\end{figure}

To validate our timing model against empirical data, we measure time per batch during end-to-end, \nwise{2}, and \nwise{3} training of Transformer models, distributed over $4$ or $6$ accelerators. The architecture is as described in Section~\ref{subsection:transformer-results}, and the results are shown in Figure~\ref{fig:e2e-experimental-timing}. The micro-batch cost parameters $c_0, c_1$ are fitted to match the timing data, as they inherently depend on the hardware and model architecture.

The ratio $c_0/c_1$ controls the speedups offered by \nwise{n} -- the higher it is, the less effective fine micro-batching is, and the more beneficial \nwise{n} is.

Finally, it is worth noting that forwards passes might be significantly faster than backwards passes, which is not modelled by our timing model. For instance, in a setting where forwards passes are two times faster than backwards passes, the speed of end-to-end learning is underestimated by the model by 25\%. However, that setting still allows Interlocking Backpropagation to achieve large step time improvements for deep enough models.


\subsection{Optimal number of micro-batches}

Assuming our timing model, the optimal number of micro-batches to use with \nwise{n} is $2$ (when $N < 1 + A/2$), while for end-to-end training, it grows as $\Theta(\sqrt{A})$. Figure~\ref{fig:nwise_speedups} assumes the optimal micro-batch size at each point of the plot, only constraining $M$ to be a power of $2$. Those optimal values are listed in Table~\ref{tab:optimal-m-values}.

\begin{table}
    \centering
    \begin{tabular}{r | c | c | c | c | c | c }

        \textbf{Number of accelerators $A$} & 2 & 3 & 4 & 5 & 6 -- 11 & 12 -- 30 \\\specialrule{1.5pt}{0pt}{0pt}
        \text{Optimal $M$ for end-to-end} & 8 & 8 & 16 & 16 & 16 & 32 \\\hline
        \text{Optimal $M$ for \nwise{1}} & 1  & 1 & 1 & 1 & 1 & 1 \\\hline
        \text{Optimal $M$ for \nwise{2}} & 8  & 2 & 2 & 2 & 2 & 2 \\\hline
        \text{Optimal $M$ for \nwise{3}} & 8  & 4 & 2 & 2 & 2 & 2 \\\hline
        \text{Optimal $M$ for \nwise{4}} & 8  & 8 & 4 & 2 & 2 & 2 \\\hline
        \text{Optimal $M$ for \nwise{5}} & 16 & 8 & 8 & 4 & 2 & 2
    \end{tabular}
    \caption{Optimal number of micro-batches (restricted to powers of $2$) per mini-batch for a range of training strategies, and increasing numbers of accelerators, given micro-batch cost parameters ${c_0=0.025\text{s}, c_1=1.279\text{s}}$.}
    \label{tab:optimal-m-values}
\end{table}

\subsection{Other forms of model-parallelism}
\label{app:other_model_parallelism}

The term `model parallelism' has been used in literature to refer to two distinct ideas -- pipeline parallelism (an example of which is GPipe, discussed in this work) and tensor parallelism, also known as tensor sharding, introduced in the Megatron framework~\citep{megatron}.
Tensor parallelism pertains to splitting large tensors, and computations involving them, between multiple accelerators.

Tensor parallelism, pipeline parallelism, and interlocking backpropagation can all be used or disabled independently for any model. The interaction of tensor parallelism and interlocking backpropagation is less complex than that of GPipe and interlocking backpropagation, as the former would simply affect the time of processing a micro-batch, $c(M)$, in our timing model, at the cost of consuming extra accelerators.

Some timing analysis -- akin to ours in its goals -- of how pipeline and tensor parallelism interact, can be found in \citet{parallelism-timing-analysis}. Together with our timing analysis, it could provide insight on how to optimally use interlocking backpropagation, GPipe, and tensor sharding all together.
Furthermore, an automated framework for model parallelism, compatible with TPU accelerators, has been recently released as GSPMD~\citep{GSPMD}, and could act as a starting point for practitioners interested in implementing interlocking backpropagation combined with other forms of model parallelism.

\end{appendices}


\end{document}